\PassOptionsToPackage{unicode}{hyperref}
\PassOptionsToPackage{hyphens}{url}
\PassOptionsToPackage{dvipsnames,svgnames,x11names}{xcolor}
\documentclass[
]{article}

\usepackage{amsmath,amssymb}
\usepackage{iftex}
\ifPDFTeX
  \usepackage[T1]{fontenc}
  \usepackage[utf8]{inputenc}
  \usepackage{textcomp} % provide euro and other symbols
\else % if luatex or xetex
  \usepackage{unicode-math}
  \defaultfontfeatures{Scale=MatchLowercase}
  \defaultfontfeatures[\rmfamily]{Ligatures=TeX,Scale=1}
\fi
\usepackage{lmodern}
\ifPDFTeX\else  
\fi
\IfFileExists{upquote.sty}{\usepackage{upquote}}{}
\IfFileExists{microtype.sty}{% use microtype if available
  \usepackage[]{microtype}
  \UseMicrotypeSet[protrusion]{basicmath} % disable protrusion for tt fonts
}{}
\makeatletter
\@ifundefined{KOMAClassName}{% if non-KOMA class
  \IfFileExists{parskip.sty}{%
    \usepackage{parskip}
  }{% else
    \setlength{\parindent}{0pt}
    \setlength{\parskip}{6pt plus 2pt minus 1pt}}
}{% if KOMA class
  \KOMAoptions{parskip=half}}
\makeatother
\usepackage{xcolor}
\ifx\paragraph\undefined\else
  \let\oldparagraph\paragraph
  \renewcommand{\paragraph}[1]{\oldparagraph{#1}\mbox{}}
\fi
\ifx\subparagraph\undefined\else
  \let\oldsubparagraph\subparagraph
  \renewcommand{\subparagraph}[1]{\oldsubparagraph{#1}\mbox{}}
\fi

\providecommand{\tightlist}{%
  \setlength{\itemsep}{0pt}\setlength{\parskip}{0pt}}\usepackage{longtable,booktabs,array}
\usepackage{calc} % for calculating minipage widths
\usepackage{etoolbox}
\makeatletter
\patchcmd\longtable{\par}{\if@noskipsec\mbox{}\fi\par}{}{}
\makeatother
\IfFileExists{footnotehyper.sty}{\usepackage{footnotehyper}}{\usepackage{footnote}}
\makesavenoteenv{longtable}
\usepackage{graphicx}
\makeatletter
\def\maxwidth{\ifdim\Gin@nat@width>\linewidth\linewidth\else\Gin@nat@width\fi}
\def\maxheight{\ifdim\Gin@nat@height>\textheight\textheight\else\Gin@nat@height\fi}
\makeatother
\setkeys{Gin}{width=\maxwidth,height=\maxheight,keepaspectratio}
\makeatletter
\def\fps@figure{htbp}
\makeatother
\newlength{\cslhangindent}
\newlength{\csllabelwidth}
\newlength{\cslentryspacingunit} % times entry-spacing
\newenvironment{CSLReferences}[2] % #1 hanging-ident, #2 entry spacing
 {% don't indent paragraphs
  \setlength{\parindent}{0pt}
  \ifodd #1
  \let\oldpar\par
  \def\par{\hangindent=\cslhangindent\oldpar}
  \fi
  \setlength{\parskip}{#2\cslentryspacingunit}
 }%
 {}
\usepackage{calc}

\usepackage{booktabs}
\usepackage{longtable}
\usepackage{array}
\usepackage{multirow}
\usepackage{wrapfig}
\usepackage{float}
\usepackage{colortbl}
\usepackage{pdflscape}
\usepackage{tabu}
\usepackage{threeparttable}
\usepackage{threeparttablex}
\usepackage[normalem]{ulem}
\usepackage{makecell}
\usepackage{xcolor}
\usepackage{multicol}
\usepackage{hhline}
\usepackage{hyperref}
\usepackage{arxiv}
\usepackage{orcidlink}
\usepackage{amsmath}
\usepackage[T1]{fontenc}
\usepackage{amssymb,amsthm}
\usepackage{physics}
\usepackage{tabularx}

\newtheorem{definition}{Definition}

\DeclareMathOperator*{\argmin}{argmin}

\newcommand{\RR}{\mathbb{R}}  % Real space

\newcommand{\length}[1]{\vert#1\vert}

\newcommand{\StateSpace}{\mathbb{S}}  % Space of states
\newcommand{\ObsSpace}{\mathbb{O}}  % Space of observations
\newcommand{\ActionSpace}{\mathbb{A}}  % Space of actions
\newcommand{\Q}{\texttt{Q}}  % Q function
\newcommand{\V}{\texttt{V}}  % Value function
\newcommand{\pos}{\texttt{P}}  % Position function (in a SOM)
\newcommand{\RewardFn}{\texttt{R}}
\newcommand{\LAgts}{\mathcal{L}}  % Set of learning agents

\makeatletter
\@ifpackageloaded{caption}{}{\usepackage{caption}}
\AtBeginDocument{%
\ifdefined\contentsname
  \renewcommand*\contentsname{Table of contents}
\else
  \newcommand\contentsname{Table of contents}
\fi
\ifdefined\listfigurename
  \renewcommand*\listfigurename{List of Figures}
\else
  \newcommand\listfigurename{List of Figures}
\fi
\ifdefined\listtablename
  \renewcommand*\listtablename{List of Tables}
\else
  \newcommand\listtablename{List of Tables}
\fi
\ifdefined\figurename
  \renewcommand*\figurename{Figure}
\else
  \newcommand\figurename{Figure}
\fi
\ifdefined\tablename
  \renewcommand*\tablename{Table}
\else
  \newcommand\tablename{Table}
\fi
}
\@ifpackageloaded{float}{}{\usepackage{float}}
\floatstyle{ruled}
\@ifundefined{c@chapter}{\newfloat{codelisting}{h}{lop}}{\newfloat{codelisting}{h}{lop}[chapter]}
\floatname{codelisting}{Listing}

\usepackage{amsthm}
\theoremstyle{remark}
\AtBeginDocument{}
\newtheorem*{remark}{Remark}

\makeatother
\makeatletter
\@ifpackageloaded{caption}{}{\usepackage{caption}}
\@ifpackageloaded{subcaption}{}{\usepackage{subcaption}}
\makeatother
\makeatletter
\@ifpackageloaded{tcolorbox}{}{\usepackage[skins,breakable]{tcolorbox}}
\makeatother
\makeatletter
\@ifundefined{shadecolor}{\definecolor{shadecolor}{rgb}{.97, .97, .97}}
\makeatother
\makeatletter
\@ifpackageloaded{algorithm}{}{\usepackage{algorithm}}
\makeatother
\makeatletter
\@ifpackageloaded{algpseudocode}{}{\usepackage{algpseudocode}}
\makeatother
\ifLuaTeX
  \usepackage{selnolig}  % disable illegal ligatures
\fi
\IfFileExists{bookmark.sty}{\usepackage{bookmark}}{\usepackage{hyperref}}
\IfFileExists{xurl.sty}{\usepackage{xurl}}{} % add URL line breaks if available
\hypersetup{
  pdftitle={Adaptive reinforcement learning of multi-agent ethically-aligned behaviours: the QSOM and QDSOM algorithms},
  pdfauthor={Remy Chaput; Olivier Boissier; Mathieu Guillermin},
  pdfkeywords={Machine Ethics, Artificial Moral Agents, Multi-Agent
Systems, Reinforcement Learning, Multi-Agent Reinforcement Learning},
  colorlinks=true,
  linkcolor={blue},
  filecolor={Maroon},
  citecolor={Blue},
  urlcolor={Blue},
  pdfcreator={LaTeX via pandoc}}

\newcommand{\runninghead}{A Preprint }
\renewcommand{\runninghead}{A Preprint }
\title{Adaptive reinforcement learning of multi-agent ethically-aligned
behaviours: the QSOM and QDSOM algorithms}
\author{
\textbf{Remy Chaput}~\orcidlink{0000-0002-2233-7566}\\Univ Lyon, UCBL,
CNRS, INSA Lyon, Centrale Lyon, Univ Lyon 2, LIRIS, UMR5205\\F-69622,
Villeurbanne\\France\\\href{mailto:remy.chaput@univ-lyon1.fr}{remy.chaput@univ-lyon1.fr}\\\\\\
\textbf{Olivier Boissier}\\Institut Henri Fayol\\MINES
Saint-Etienne\\France\\\href{mailto:Olivier.Boissier@emse.fr}{Olivier.Boissier@emse.fr}\\\\\\
\textbf{Mathieu Guillermin}\\CONFLUENCE: Sciences et Humanités research
unit (EA 1598)\\Lyon Catholic
University\\France\\\href{mailto:mguillermin@univ-catholyon.fr}{mguillermin@univ-catholyon.fr}}
\date{}
\begin{document}
\maketitle
\begin{abstract}
The numerous deployed Artificial Intelligence systems need to be aligned
with our ethical considerations. However, such ethical considerations
might change as time passes: our society is not fixed, and our social
mores evolve. This makes it difficult for these AI systems; in the
Machine Ethics field especially, it has remained an under-studied
challenge. In this paper, we present two algorithms, named QSOM and
QDSOM, which are able to adapt to changes in the environment, and
especially in the reward function, which represents the ethical
considerations that we want these systems to be aligned with. They
associate the well-known Q-Table to (Dynamic) Self-Organizing Maps to
handle the continuous and multi-dimensional state and action spaces. We
evaluate them on a use-case of multi-agent energy repartition within a
small Smart Grid neighborhood, and prove their ability to adapt, and
their higher performance compared to baseline Reinforcement Learning
algorithms.
\end{abstract}
{\bfseries \emph Keywords}
\def\sep{\textbullet\ }
Machine Ethics \sep Artificial Moral Agents \sep Multi-Agent
Systems \sep Reinforcement Learning \sep 
Multi-Agent Reinforcement Learning

\ifdefined\Shaded\renewenvironment{Shaded}{\begin{tcolorbox}[enhanced, interior hidden, borderline west={3pt}{0pt}{shadecolor}, sharp corners, boxrule=0pt, frame hidden, breakable]}{\end{tcolorbox}}\fi

\floatname{algorithm}{Algorithm}

\floatname{algorithm}{Algorithm}

\hypertarget{sec-intro}{%
\section{Introduction}\label{sec-intro}}

With the increasing deployment of systems using Artificial Intelligence
(AI) techniques, questions are being raised within civil society and the
scientific community about their impact on our lives. One of the most
pressing questions is that of value alignment (Dignum 2019 ; World
Economic Forum 2015): how can we ensure that these systems act in line
with the moral values that are important to us? The field of Machine
Ethics has proposed numerous approaches, based on a variety of
techniques, from symbolic implementation to machine learning. However,
the property of Continual Learning, which we believe is important, has
not been studied enough. Continual Learning concerns the ability of
artificial agents to learn continuously and therefore to change their
behaviour as a function of the environment. This is a particularly
critical property in Machine Ethics, because ethics are not fixed: our
currently accepted social mores evolve over time. In this paper, we
propose in Section~\ref{sec-model} two reinforcement learning
algorithms, QSOM and QDSOM, that can adapt to changes in the reward
function, representing these ``changes in ethics''. These algorithms are
then evaluated on an application case of multi-agent energy repartition
within a small Smart Grid, described in Section~\ref{sec-experiments}. A
discussion of their advantages and drawbacks is finally presented in
Section~\ref{sec-discussion}.

\hypertarget{sec-sota}{%
\section{State of the Art}\label{sec-sota}}

In this section, we introduce the necessary knowledge, and explore the
state of the art in the fields related to our work: \emph{Machine
Ethics} and \emph{(Multi-Agent) Reinforcement Learning}. This
exploration allows us to compare the existing approaches, their
advantages, but also their limitations, and to define some concepts
necessary to the understanding of our work.

\hypertarget{sota-machineethics}{%
\subsection{Machine Ethics}\label{sota-machineethics}}

The field of Machine Ethics is relatively recent among the other fields
of Artificial Intelligence. A book published in 2011 gathers different
essays on the nature of \emph{Machine Ethics}, its importance, the
difficulties and challenges to be solved, and also a few first
approaches (Anderson and Anderson 2011). This book defines this new
field of research:

\begin{quote}
The new field of machine ethics is concerned with giving machines
ethical principles, or a procedure for discovering a way to resolve the
ethical dilemmas we might encounter, enabling them to function in an
ethically responsible manner through their own ethical decision making.
(Anderson and Anderson 2011)
\end{quote}

Being a recent field, several articles have sought to position
themselves, or to offer a philosophical background. For example, Moor
(2009) proposes a definition of what might be an ``ethical robot'', and
differentiates 4 different kinds of robots, ranging from those with the
least ethical considerations to those which have near-human ethical
reasoning abilities: \emph{ethical impact agents}, \emph{implicit
ethical agents}, \emph{explicit ethical agents}, and \emph{full ethical
agents}. The goal, for Machine Ethics designers and researchers, is to
attain \emph{explicit} ethical agents, as it is still unsure whether
artificial \emph{full} ethical agents can be built.

In the following, we briefly list a few approaches, and present a set of
``properties'' that we argue are important to design such ethical
agents.

\textbf{Discrete or continuous domains}. In order to implement ethical
considerations into an artificial agent, these considerations must be
represented. This includes, e.g., data about the current situation, and
the potential actions or decisions that are available to the agent. The
choice of this representation must allow both for use-case richness, and
for the agent's ability to correctly use these representations. Two
types of representations are commonly used: either \emph{discrete}
domains, which use a discrete set of symbols and discrete numbers, or
\emph{continuous} domains, which use continuous numbers that lead to an
infinite set of symbols.

So far, discrete domains seem prevalent in \emph{Machine Ethics}. For
example, the emblematic Trolley Dilemma (Foot 1967) describes a
situation where an uncontrolled trolley is driving on tracks towards a
group of 5 persons. These persons, depending on the exact specification,
are either unaware of the trolley, or unable to move. An agent may save
this group by pulling up a lever, which would derail the trolley towards
a single person. It can be seen that the representation of both the
situation and the available actions are discrete in this dilemma: 2
actions are proposed, \emph{pull the lever} or \emph{do nothing}, and on
the tracks are present \emph{1} and \emph{5} persons, respectively.

Similarly, the now defunct
\href{https://web.archive.org/web/20210323073233/https://imdb.uib.no/dilemmaz/articles/all}{DilemmaZ
database} listed a plethora of moral dilemmas, proposed by the
community, of which many apply to Artificial Intelligence and IT systems
in general, e.g., smart homes, robots. Although a formal description of
these dilemmas is not available, most of the natural language
descriptions seem to imply discrete features. This is particularly clear
for the definition of actions; for example, the ``Smart home - Someone
smoking marijuana in a house'' dilemma, by Louise A. Dennis, offers the
following 3 actions: ``a) do nothing, b) alert the adults and let them
handle the situation or c) alert the police''.

A final example is the \emph{Moral Gridworlds} idea of Haas (2020) to
train a Reinforcement Learning agent ``to attribute subjective rewards
and values to certain `moral' actions, states of affairs, commodities,
and perhaps even abstract representations''. Moral Gridworlds are based
on gridworlds, which represent the environment as a 2-dimensional grid
of cells. A RL agent is placed in one of these cells, and may either act
in its cell, or move to one of the adjacent cells. Again, the
environment uses discrete features, both for perception, i.e., a
discrete set of cells, and for actions, i.e., either act, move up, left,
right, or down.

Perhaps the ubiquitous use of discrete representations in Machine Ethics
can be at least partially explained by their simplicity of usage within
AI techniques. These ``discrete dilemmas'' are important, because they
may very well happen one day in our society. We need systems that are
able to make the best decision, with respect to our moral values, in
such situations.

However, there are other situations that cannot be easily described by
discrete representations. For example, foretelling the Smart Grid
use-case that we describe in Section~\ref{sec-experiments}, when
considering an energy distribution system, we may transition from a
closed question ``Should the agent consume energy? yes/no'' to a more
open question ``What power should the agent request during a given time
step?''. Arguably, such an action could be represented as a discrete
set, by \emph{discretizing} the continuous domain into a set, e.g.,
\(\left\{ 0\textrm{Wh}, 1\textrm{Wh}, \cdots, 1000\textrm{Wh} \right\}\),
which contains \(1001\) actions. But this solution is harder to leverage
when considering multi-dimensional domains: in addition to ``how much
energy should it consume'', we may also ask ``What power should the
agent buy?''. In this case, discretizing the continuous and
multi-dimensional domain would result in a combinatorial explosion. The
set of discrete actions may be represented as
\(\left\{ (0\textrm{Wh}, 0\textrm{Wh}), (0\textrm{Wh}, 1\textrm{Wh}), (1\textrm{Wh}, 0\textrm{Wh}), (1\textrm{Wh}, 1\textrm{Wh}), \cdots, (1000\textrm{Wh}, 1000\textrm{Wh}) \right\}\),
which contains \(1001 \times 1001\) different actions, where each action
is represented as a pair \((\mathrm{consumed}, \mathrm{bought})\). We
already see, on 2 dimensions and with a grain of \(1\)Wh, that a million
actions would require too much time and computational resources to
explore and analyze, in order to find the best one. The same argument
can be made for perceptions as well: for example, instead of having a
perception ``the situation is fair'', or ``the situation is unfair'', we
may want to have an indicator of how fair the situation is, e.g.,
through well-known measures such as the Gini index, which is a real
number comprised between \(0\) (perfect equality) and \(1\) (perfect
inequality) (Gini 1936).

Such situations, which imply a large, continuous and multi-dimensional
domain, are as likely to happen in our society as the discrete ones.

\textbf{Mono- or Multi-agent}.

According to a survey (Yu et al. 2018), many works consider a single
agent isolated in its environment. This is the case, to give some
examples, of GenEth (Anderson, Anderson, and Berenz 2018), or the
\emph{ethics shaping} technique (Wu and Lin 2018). Other approaches,
such as Ethicaa (Cointe, Bonnet, and Boissier 2016), use multiple
agents, which take actions and have an impact in a common, shared
environment.

As Murukannaiah et al. (2020) put it:

\begin{quote}
Ethics is inherently a multiagent concern --- an amalgam of (1) one
party's concern for another and (2) a notion of justice.
\end{quote}

In Ethicaa (Cointe, Bonnet, and Boissier 2016), a judgment process is
defined to allow agents to both 1) select the best ethical action that
they should make, and 2) judge the behaviour of other agents so as to
determine whether they can be deemed as ``ethical'', with respect to
one's own preferences and upheld moral values. One long-term objective
of this second point can be to define and compute a trust indicator for
other agents; if an agent acts ethically, we may trust it. This raises
an interesting rationale for exploring Machine Ethics in Multi-Agent
Systems: even if we manage to somehow create a full ethical agent, which
is guaranteed to take moral values and ethical stakes into account, it
will have to work with other agents. We cannot guarantee that these
agents will follow the same ethical preferences, nor even that they will
consider ethical stakes at all. Our own agent must therefore take this
into account.

Based on the previous reasons, we argue that the multi-agent case is
important. Indeed, it corresponds to a more realistic situation: such
artificial agents are bound to be included in our society, and thus to
have to interact with other agents, whether artificial or human, or at
least to live in an environment impacted by these other agents, and not
in a perfectly isolated world. The question of the impact of other
agents on an agent's decision-making is thus of primary importance.

\textbf{Top-Down, Bottom-Up, and Hybrid approaches}.

Approach type is probably the most discussed property in \emph{Machine
Ethics}. It characterizes the way designers implement ethical
considerations into artificial agents. Similarly to the usual
classification in AI, works are divided into 3 categories (Allen, Smit,
and Wallach 2005): \emph{Top-Down}, \emph{Bottom-Up}, and \emph{Hybrid}
approaches.

Top-Down approaches are interested in formalizing existing ethical
principles from moral philosophy, such as Kant's Categorical Imperative,
or Aquinas' Doctrine of Double Effect. The underlying idea is that, if
these moral theories could be transformed into an algorithm that agents
could follow to the letter, surely these agents' behaviour would be
deemed as ethical by human observers.

This formalization is often done through symbolic representation and
reasoning, e.g., through logic, rules-based techniques, or even
ontologies. Reasoning over these symbolic representations can rely upon
expert knowledge, a priori injected. They also offer a better
readability, of both the injected knowledge, and the resulting
behaviour.

One of the advantages of Top-Down approaches is this ability to leverage
such existing ethical principles from moral philosophy. Intuitively, it
seems indeed better to rely on theories proposed by moral philosophers,
which have been tested and improved over time.

Another advantage, emphasized by the work of Bremner et al. (2019), is
the ability to use formal verification to ensure that agents' behaviours
stay within the limits of the specified rules. To do so, the
\emph{Ethical Layer} they propose includes a planning module that
creates plans, i.e., sequences of actions, and an ethical decision
module to evaluate the plans, prevent unethical ones, and proactively
ask for new plans if necessary. This formal verification ability is an
important strength, as there are worries about agents malfunctioning. An
agent that could be formally verified to stay within its bounds, could
be said to be ``ethical'', with respect to the chosen ethical principle
or theory.

However, there are some weaknesses to \emph{Top-Down} approaches. For
example, conflicts between different rules may arise: a simple conflict
could be, for example, between the ``Thou shalt not kill'' rule, and
another ``You may kill only to defend yourself''. The second one should
clearly define when it is allowed to take precedence over the first one.
A more complicated conflict would be two rules that commend different,
non-compatible actions. For example, let us imagine two missiles
attacking two different buildings in our country: the first one is a
hospital, the second one is a strategic, military building, hosting our
defense tools. An autonomous drone can intercept and destroy one of the
two missiles, but not the two of them; which one should be chosen? A
rule may tell us to protect human lives, whereas another encourages us
to defend our arsenal, in order to be able to continue protecting our
country. These two rules are not intrinsically in conflict, unlike our
previous example: we would like to follow both of them, and to destroy
the two missiles. Unfortunately, we are physically constrained, and we
must make a choice. Thus, a rule has to be preferred to the other.

\emph{Ethicaa} (Cointe, Bonnet, and Boissier 2016) agents make a
distinction between the moral values and ethical principles, and they
consider multiple ethical principles. Each ethical principle determines
whether an action is ethical, based on the permissible and moral
evaluations. Multiple actions can thus be evaluated as ethical by the
ethical principles, and, in many cases, there is no single action
satisfying all ethical principles. To solve this issue, agents also
include a priority order over the set of ethical principles known to
them. In this way, after an agent determines the possible, moral, and
ethical actions, it can choose an action, even if some of its rules
disagree and commend different actions. To do so, they filter out the
actions that are not evaluated as ethical, and thus should not be
selected, by their most preferred ethical principle, according to the
ethical priority order. As long as multiple actions remain considered,
they move on to the next preferred ethical principle, and so on, until a
single action remains.

Finally, another drawback is the lack of adaptability of these
approaches. Indeed, due to their explicit but fixed knowledge base, they
cannot adapt to an unknown situation, or to an evolution of the ethical
consensus within the society. We argue that this capability to adapt is
particularly important. It is similar to what Nallur (2020) calls the
\emph{Continuous Learning} property:

\begin{quote}
Any autonomous system that is long-lived must adapt itself to the humans
it interacts with. All social mores are subject to change, and what is
considered ethical behaviour may itself change.
\end{quote}

We further note that, in his landscape, only 1 out of 10 considered
approaches possesses this ability (Nallur 2020, Table 2).

Bottom-Up approaches try to learn a behaviour through experience, e.g.,
from a dataset of labeled samples, or trial and error interactions.

For example, GenEth (Anderson, Anderson, and Berenz 2018) uses
ethicists' decisions in multiple situations as a dataset representing
the ethical considerations that should be embedded in the agent. This
dataset is leveraged through Inductive Logic Programming (ILP) to learn
a logical formula that effectively drives the agent's behaviour, by
determining the action to be taken in each situation. ILP allows
creating a logical formula sufficiently generic to be applied to other
situations, not encountered in the dataset. An advantage of this
approach is that it learns directly from ethicists' decisions, without
having to program it by hand. The resulting formula may potentially be
understandable, provided that it is not too complex, e.g., composed of
too many terms or terms that in themselves are difficult to understand.

Another approach proposes to use Reinforcement Learning RL (Wu and Lin
2018). Reinforcement Learning relies on rewards to reinforce, or on
contrary, to mitigate a given behaviour. Traditionally, rewards are
computed based on the task we wish to solve. In the work of Wu and Lin
(2018), an ethical component is added to the reward, in the form of a
difference between the agent's behaviour, and the behaviour of an
average human, obtained through a dataset of behaviours, and supposedly
exhibiting ethical considerations. The final reward, which is sent to
agents, is computed as the sum of the ``task'' reward, and the
``ethical'' reward. Agents thus learn to solve their task, while
exhibiting the ethical considerations that are encoded in the human
samples. One advantage of this approach is that the ``ethical'' part of
the behaviour is mostly task-agnostic. Thus, only the task-specific
component of the reward has to be crafted by designers for a new task.
Nevertheless, one may wonder to which extent does this dataset really
exhibit ethical considerations? We humans do not always respect laws or
moral values, e.g., we sometimes drive too fast, risking others' lives,
or we act out of spite, jealousy, etc. To determine whether this dataset
is appropriate, an external observer, e.g., a regulator, an ethicist, or
even a concerned citizen, has to look at its content, and understand the
data points.

These 2 approaches, although based on learning, have not considered the
question of long-term adaptation to changing situations and ethical
mores. Indeed, if the current society norms with regard to ethics
change, these agents' behaviours will have to change as well. It will
probably require to create a new dataset, and to learn the agents again,
from scratch, on these new data.

Moreover, Bottom-Up approaches are harder to interpret than Top-Down
ones. For example, a human regulator or observer, willing to understand
the expected behaviour, will have to look at the dataset, which might be
a tedious task and difficult to apprehend, because of both its structure
and the quantity of data. This is all the more true with Deep Learning
approaches, which require an enormous amount of data (Marcus 2018),
making datasets exploration even more daunting.

Finally, Hybrid approaches combine both Top-Down and Bottom-Up, such
that agents are able to learn ethical behaviours by experience, while
being guided by an existing ethical framework to enforce constraints and
prevent them from diverging. As Dignum (2019) points out:

\begin{quote}
By definition, hybrid approaches have the potential to exploit the
positive aspects of the top-down and bottom-up approaches while avoiding
their problems. As such, these may give a suitable way forward. (Dignum
2019, 81)
\end{quote}

One of such hybrid works is the approach by Honarvar and Ghasem-Aghaee
(2009) to combine BDI agents with Case-based Reasoning and an Artificial
Neural Network. Faced with a given situation, the agent proposes an
action to perform, and then searches its database of already known cases
for similar situations and similar actions. If a close enough case is
found, and the action was considered as ethical in this case, the action
is taken. However, if in this close enough case, the action was
considered as unethical, a new action is requested, and the agent
repeats the same algorithm. If the agent does not have a sufficiently
close case, it performs the action, and uses its neural network to
evaluate the action's consequences and determine whether it was
effectively aligned with the ethical considerations. This evaluation is
memorized in the case database, to be potentially reused during the next
decision step. This approach indeed combines both reasoning and learning
capabilities; however, it may be difficult to apply. Case-based
reasoning allows grouping close situations and actions, but requires to
specify how to group them, i.e., what is the distance function, and how
to adapt an evaluation when either the situation or the action differs.
For example, let us assume that, in a situation \(s\), the agent's
action was to consume \(500\)Wh of energy, and the action was evaluated
as ethical. In a new situation, \(s'\), which is deemed as similar to
\(s\) by the case-based reasoner, another action is proposed, which is
to consume \(600\)Wh. Is this action ethical? How can we translate the
difference between \(600\) and \(500\) in terms of ethical impact? This
requires specifying an ``adaptation knowledge'' that provides the
necessary knowledge and tools.

Still, Hybrid approaches offer the possibility of learning a behaviour,
thus adapting to any change in the environment, while still guiding or
constraining the agent through symbolic reasoning and knowledge, thus
injecting domain expert knowledge, more easily understandable and
modifiable than datasets of examples.

\hypertarget{sota-rl}{%
\subsection{Reinforcement Learning}\label{sota-rl}}

We propose to use Reinforcement Learning (RL) as a method to learn
behaviours aligned with moral values, and provide here the background
knowledge and concepts that are necessary to understand the rest of the
paper. We detail motivations for using RL, definitions of core concepts,
and equations.

RL is a method to learn a behaviour, mainly by using trial-and-error.
Sutton and Barto (2018) define it as follows:

\begin{quote}
Reinforcement learning problems involve learning what to do --- how to
map situations to actions --- so as to maximize a numerical reward
signal. (Sutton and Barto 2018, 2)
\end{quote}

To do so, learning agents are placed in a closed-loop with an
environment, with which they interact. Through the environment, they
have knowledge of which state they are in, and they take actions to
change the state. One of the key points of RL is that learning agents
are not told which action is the correct one; the feedback they receive,
or \emph{reward}, merely tells them to which degree the action was
satisfying. Learning agents must discover the best action, i.e., the one
that yields the highest reward, by accumulating enough experience, that
is by repetitively trying each action in each situation, and observing
the received rewards.

As we mentioned, RL agents receive feedback, which differentiates them
from the \emph{unsupervised} paradigm. However, unlike the
\emph{supervised} paradigm, this feedback does not clearly indicate
which was the correct answer. This removes the assumption that we know
the correct answer to each input. Instead, we provide a reward function,
and thus optimize the agent's output step by step, by improving the
proposed action based on the reward.

The goal of a RL algorithm is to learn a policy, or strategy, denoted
\(\pi\), such that the agent knows which action to take in each
situation. \(\pi\) is often defined as
\(\pi : \StateSpace \to \ActionSpace\) in the case of a
\emph{deterministic} policy, where \(\StateSpace\) is the space of
possible states, and \(\ActionSpace\) the space of possible actions. To
each state \(s\) is associated a single action \(\pi(s) = a\), which the
agent should take in order to maximize its reward. Another formulation
is \(\pi : \StateSpace \times \ActionSpace \to [0,1]\), in the case of a
\emph{stochastic} policy. For each combination of state-action \((s,a)\)
is associated a probability \(\pi(s,a)\) of taking action \(a\) in the
state \(s\), such that
\(\forall s \in \StateSpace : \sum_{\forall a \in \ActionSpace} \pi(s,a) = 1\).

There are several challenges in RL, of which one of the most known and
perhaps important is the \emph{exploration-exploitation trade-off}. In
order to facilitate learning the policy function, RL researchers often
rely on the notion of \emph{values}\footnote{We use value here in a
  different sense than the moral value used earlier. To avoid confusion,
  we will always specify \emph{moral} value when referring to this first
  meaning.}, in aptly-named \emph{value-based} methods, such as the
well-known Q-Learning (Watkins and Dayan 1992). The value of a state, or
a state-action pair, represents the long-term interest of being in this
state, whereas the reward is short-term feedback. The agent could
receive a high reward for taking an action \(a\) in a state \(s\), but
ending up in a state \(s'\) in which only low rewards can be obtained.
In this case, we will say that the value of state \(s'\), denoted as
\(\V(s')\) is low. By extension, the agent has little interest in
performing action \(a\) while in state \(s\), since it will lead it to a
low-interest state.

In the previous paragraph, we derived the interest of action \(a\), in a
state \(s\), from the value \(\V(s')\) which it leads to. It is also
possible to learn directly the value of state-action pairs, which is the
main idea of the Q-Learning algorithm. To retain the different interests
of all state-action pairs, a table, named the \emph{Q-Table}, is
created, having the states as columns and actions as rows. The
\emph{Q-Value} \(\Q(s,a)\) is thus defined as the interest of the
state-action pair \((s,a)\), i.e., the interest of taking action \(a\)
in state \(s\). Additionally, the value of a state as a whole is defined
as \(\V(s) = max_{a} \Q(s,a)\).

Based on these definitions, the agent is able to learn the Q-Table by
iteratively collecting experiences from the environment, in the form of
\(\left\langle s, a, s', r \right\rangle\) tuples, updating the interest
\(\Q(s,a)\) based on both the short-term reward \(r\), and the long-term
interest \(\V(s')\) of arriving in state \(s'\). Mathematically, this
can be solved through dynamic programming, by applying the Bellman
equation on the Q-Values (Bellman 1966):

\begin{equation}\protect\hypertarget{eq-bellman}{}{
  \Q_{t+1}(s_t,a_t) \leftarrow \alpha \left[r_t + \gamma \max_{a'} \Q_{t}(s_{t+1},a') \right] + (1 - \alpha)\Q_{t}(s_t,a_t)
}\label{eq-bellman}\end{equation}

Where \(r_t\) was the reward received at step \(t\), \(s_t\) was the
state at step \(t\), \(a_t\) was the action chosen by the agent, and
\(s_{t+1}\) is the new state resulting from performing \(a_t\) in
\(s_t\).

As the values are updated by taking the difference between the old value
and a new value, this type of methods is named the \emph{Temporal
Difference} learning, or TD-Learning.

\hypertarget{sota-marl}{%
\subsection{Multi-Agent Reinforcement Learning}\label{sota-marl}}

Although Reinforcement Learning was originally concerned with the
learning of a single agent, there are numerous cases where a multi-agent
system can, or must, be considered.

For example, let us consider a virtual agent dedicated to helping a
human user in its day-to-day tasks, such as booking appointments. The
diversity of human users implies a diversity of virtual agents, which
will have to communicate and interact together, in order to solve the
tasks of their users. In this example, the multiplicity of agents is a
necessity that stems from the social system in which we live.

One of the most important challenges that additionnally arises in
multi-agent systems is the problem of ``Multi-Agent Credit Assignment
Problem'' (MA-CAP). Several definitions of the MA-CAP have been given in
the literature, which are all very similar. We particularly appreciate
the formulation of Yliniemi and Tumer (2014) :

\begin{quote}
Each agent seeks to maximize its own reward; with a properly designed
reward signal, the whole system will attain desirable behaviors. This is
the science of credit assignment: determining the contribution each
agent had to the system as a whole. Clearly quantifying this
contribution on a per-agent level is essential to multiagent learning.
(Yliniemi and Tumer 2014, 2)
\end{quote}

The survey of Panait and Luke (2005, 8) summarizes several methods to
assign rewards.

The \emph{Global reward} approach considers the contribution of the
whole team. Usually, the same reward is given to all agents, either by
taking the sum of contributions, or by dividing the sum of contributions
by the number of learners. In any case, a consequence is that \emph{all}
learners' rewards depend on each agent. When an agent's contribution
decreases (resp. increases), all learners see their reward decrease as
well (resp. increase). This is a simple approach that intuitively
fosters collaboration, since all agents need to perform well in order to
receive a high reward.

However, this approach does not send accurate feedback to the learners.
Let us consider a situation in which most agents have exhibited a good
behaviour, although another one has failed to learn correctly, and has
exhibited a rather bad (or uninteresting) behaviour. As the individual
reward depends on the team's efforts, the ``bad'' agent will still
receive a praising reward. It will therefore have little incentive to
change its behaviour. On the contrary, the ``good'' agents could have
received a higher reward, if it were not for their ``bad'' colleague.
Their behaviour does not necessarily need to change, however they will
still try to improve it, since they expect to improve their received
rewards.

At the opposite extreme, the \emph{Local reward} approach considers
solely the contribution of an individual agent to determine its reward.
For example, if the agents' task is to take waste to the bin, an agent's
reward will be the number of waste products that this specific agent
brought. An advantage of this approach is to discourage laziness, as the
agent cannot rely upon others to effectively achieve the task. By
definition, agents receive a feedback that is truer to to their actual
contribution.

A problem of \emph{local rewards} is that they incentivize greedy
behaviours and do not always foster collaboration. Indeed, as agents are
rewarded based on their own contribution, without taking the others into
account, they have no reason to help other agents, or even to let them
do their task. In the waste example, an agent could develop a stealing
behaviour to take out more waste products. Another common example is the
one of a narrow bridge that two agents must cross to achieve their task.
They both arrive at the bridge at the same time, and none of them is
willing to let the other one cross first, since that would reduce their
own reward, or, phrased differently, would prevent them from getting an
even higher reward. Thus, they are both stuck in a non-interesting
situation, both in the collective and individual sense, due to their
maximizing of the individual interest only.

Another method to determine an agent's contribution to the team is to
imagine an environment in which the agent had not acted. This method is
sometimes called \emph{Difference Rewards} (Yliniemi and Tumer 2014).
The idea of this method is to reward agents if their contribution was
helpful for the team, and to force a high impact of an agent's action on
its own reward. It is computed as follows:

\begin{equation}\protect\hypertarget{eq-difference-rewards}{}{
  \texttt{D}_i(z) = \texttt{G}(z) - \texttt{G}(z_{-i})
}\label{eq-difference-rewards}\end{equation}

where \(\texttt{D}_i(z)\) is the reward of an agent \(i\), based on the
context \(z\), which is both the state and the joint-action of all
agents in the environment; \(\texttt{G}(z)\) is the global reward for
the context \(z\), and \(\texttt{G}(z_{-i})\) is an hypothetical reward,
which would have been given to the team, if the agent \(i\) had not
acted in the environment. In other words, if the current environment is
better than the hypothetical one, this means the agent's action has
improved the environment. It should be rewarded positively so as to
reinforce its good behaviour. As \(\texttt{G}(z) > \texttt{G}(z_{-i})\),
the reward will effectively be positive. Conversely, if the current
environment is worse than the hypothetical one, this means the agent's
action has deteriorated the environment, or contributed negatively. The
agent should therefore receive a negative reward, or punishment, in
order to improve its behaviour. In this case, as
\(\texttt{G}(z) < \texttt{G}(z_{-i})\), the result will be negative. If
the agent did not contribute much, its reward will be low, to encourage
it to participate more, although without impairing the team's effort, as
in the bridge example. Otherwise, the global reward \(\texttt{G}(z)\)
would diminish, and the agent's reward would therefore decrease as well.
Finally, it can be noted that the other agents' actions have a low
impact on an agent reward.

\hypertarget{sec-model}{%
\section{The QSOM and QDSOM algorithms}\label{sec-model}}

As stated in the State of the Art, the algorithms that we propose need
to handle continuous and multi-dimensional state-action spaces. They are
based on an existing work (Smith 2002a, 2002b) that we extend and
evaluate in a more complex use-case. Smith's initial work proposed, in
order to handle such domains, to associate Self-Organizing Maps (SOMs)
to a Q-Table.

We first briefly explain what is a Q-Table from the Q-Learning
algorithm, and its limitations. We then present Self-Organizing Maps,
the Dynamic Self-Organizing Map variation, and how we can use them to
solve the Q-Table's limitations. We combine these components to propose
an extension of Smith's algorithm that we name \emph{QSOM}, which
leverages a Q-Table and Self-Organizing Maps (SOMs), and a new algorithm
named \emph{QDSOM}, which leverages a Q-Table and Dynamic SOMs (DSOMs).

\begin{figure}[htbp]

{\centering \includegraphics[width=0.95\textwidth,height=\textheight]{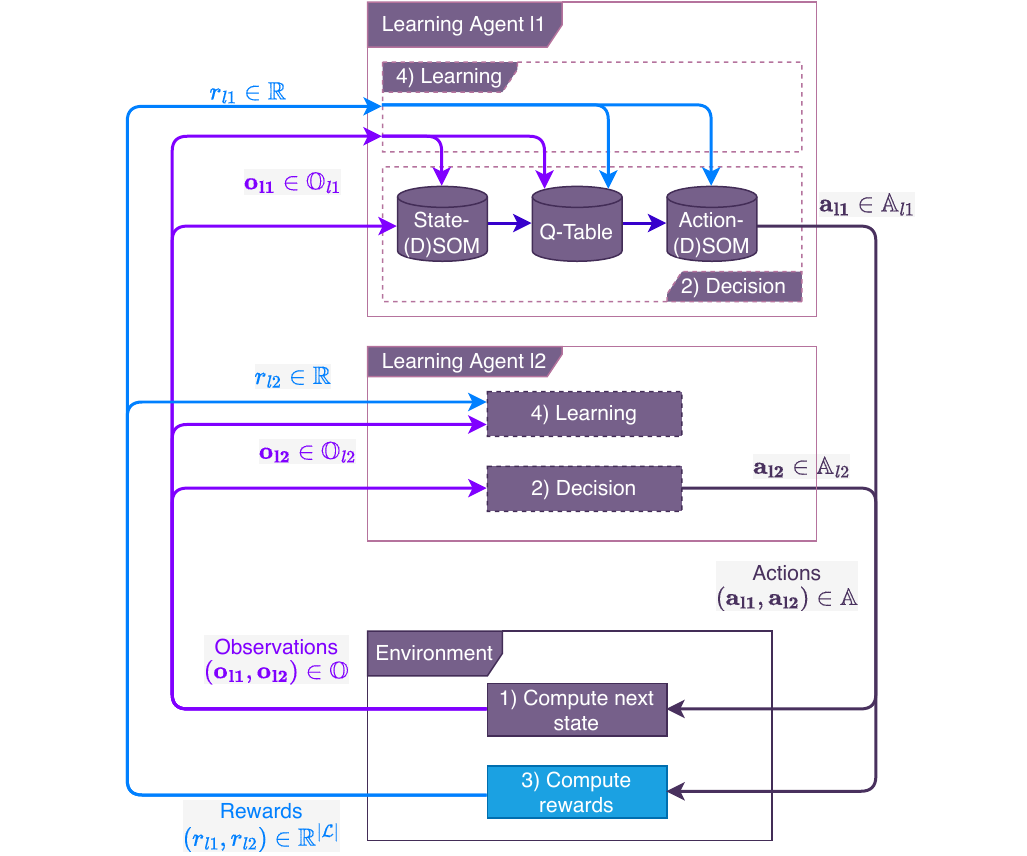}

}

\caption{\label{fig-learning1-schema}Architecture of the QSOM and QDSOM
algorithms, which consist of a decision and learning processes. The
processes rely on a State-(D)SOM, an Action-(D)SOM, and a Q-Table.}

\end{figure}

Figure~\ref{fig-learning1-schema} presents a summarizing schema of our
proposed algorithms. It includes multiple learning agents that live
within a shared environment. This environment sends observations to
agents, which represent the current state, so that agents may choose an
action and perform it in the environment. In response, the environment
changes its state, and sends them new observations, potentially
different for each agent, corresponding to this new state, as well as a
reward indicating how correct the performed action was. Learning agents
leverage the new observations and the reward to update their internal
model. This observation-action-reward cycle is then repeated so as to
make learning agents improve their behaviour, with respect to the
considerations embedded in the reward function. The decision process
relies on 3 structures, a State (Dynamic) Self-Organizing Map, also
named the State-(D)SOM, an Action (Dynamic) Self-Organizing Map, also
named the Action-(D)SOM, and a Q-Table. They take observations as inputs
and output an action, which are both vectors of continuous numbers. The
learning process updates these same structures, and takes the reward as
an input, in addition to observations.

\hypertarget{sec-learning-qtable}{%
\subsection{Q-Table}\label{sec-learning-qtable}}

The \emph{Q-Table} is the central component of the well-known Q-Learning
algorithm (Watkins and Dayan 1992). It is tasked with learning the
\emph{interest} of a state-action pair, i.e., the expected horizon of
received rewards for taking an action in a given state. The Q-Table is a
tabular structure, where rows correspond to possible states, and columns
to possible actions, such that the row \(\Q(s, \cdot)\) gives the
interests of taking every possible action in state \(s\), and, more
specifically, the cell \(\Q(s, a)\) is the interest of taking action
\(a\) in state \(s\). These cells, also named \emph{Q-Values}, can be
learned iteratively by collecting experiences of interactions, and by
applying the Bellman equation.

We recall that the interests take into account both the short-term
immediate reward, but also the interest of the following state \(s'\),
resulting from the application of \(a\) in \(s\). Thus, an action that
leads to a state where any action yields a low reward, or in other word
an unattractive state, would have a low interest, regardless of its
immediate reward.

Assuming that the Q-Values have converged towards the ``true''
interests, the optimal policy can be easily obtained through the
Q-Table, by selecting the action with the maximum interest in each
state. By definition, this ``best action'' will lead to states with high
interests as well, thus yielding, in the long-term, the maximum expected
horizon of rewards.

An additional advantage of the Q-Table is the ability to directly have
access to the interests, in comparison to other approaches, such as
\emph{Policy Gradient}, which typically manipulate actions'
probabilities, increasing and decreasing them based on received rewards.
These interests can be conveyed to humans to support or detail the
algorithm's decision process, an advantage that could be exploited for
explainability.

Nevertheless, Q-Tables have an intrinsic limitation: they are defined as
a tabular structure. This structure works flawlessly in simple
environments, e.g., those with a few discrete states and actions. Yet,
in more complex environments, especially those that require continuous
representations of states and actions, it is not sufficient any more, as
it would require an infinite number of rows and columns, and therefore
an infinite amount of memory. Additionally, because of the continuous
domains' nature, it would be almost impossible to obtain twice the exact
same state: the cells, or Q-Values, would almost always get at most a
single interaction, which does not allow for adequate learning and
convergence towards the true interests.

To counter this disadvantage, we rely on the use of Self-Organizing Maps
(SOMs) that handle the continuous domains. The mechanisms of SOMs are
explained in the next section, and we detail how they are used in
conjunction with a Q-Table in Section~\ref{sec-learning-algorithms}.

\hypertarget{sec-learning-soms}{%
\subsection{(Dynamic) Self-Organizing Maps}\label{sec-learning-soms}}

A Self-Organizing Map (SOM) (Kohonen 1990) is an artificial neural
network that can be used for unsupervised learning of representations
for high-dimensional data. SOMs contain a fixed set of neurons,
typically arranged in a rectangular 2D grid, which are associated to a
unique identifier, e.g., neuron \#1, neuron \#2, etc., and a vector,
named the \emph{prototype vector}. Prototype vectors lie in the latent
space, which is the highly dimensional space the SOM must learn to
represent.

The goal is to learn to represent as closely as possible the
distribution of data within the latent space, based on the input data
set. To do so, prototype vectors are incrementally updated and ``moved''
towards the different regions of the latent space that contain the most
data points. Each time an input vector, or data point, is presented to
the map, the neurons compete for attention: the one with the closest
prototype vector to the input vector is named the \emph{Best Matching
Unit} (BMU). Neurons' prototypes are then updated, based on their
distance to the BMU and the input vector. By doing this, the neurons
that are the closest to the input vector are moved towards it, whereas
the farthest neurons receive little to no modification, and thus can
focus on representing different parts of the latent space.

As the number of presented data points increases, the distortion, i.e.,
the distance between each data point and its closest prototype,
diminishes. In other words, neurons' prototypes are increasingly closer
to the real (unknown) distribution of data.

When the map is sufficiently learned, it can be used to perform a
mapping of high dimensional data points into a space of lower dimension.
Each neuron represents the data points that are closest to its prototype
vector. Conversely, each data point is represented by the neuron whose
prototype is the closest to its own vector.

This property of SOMs allows us to handle continuous, and
multi-dimensional state and action spaces.

\begin{figure}[htbp]

{\centering \includegraphics[width=1\textwidth,height=\textheight]{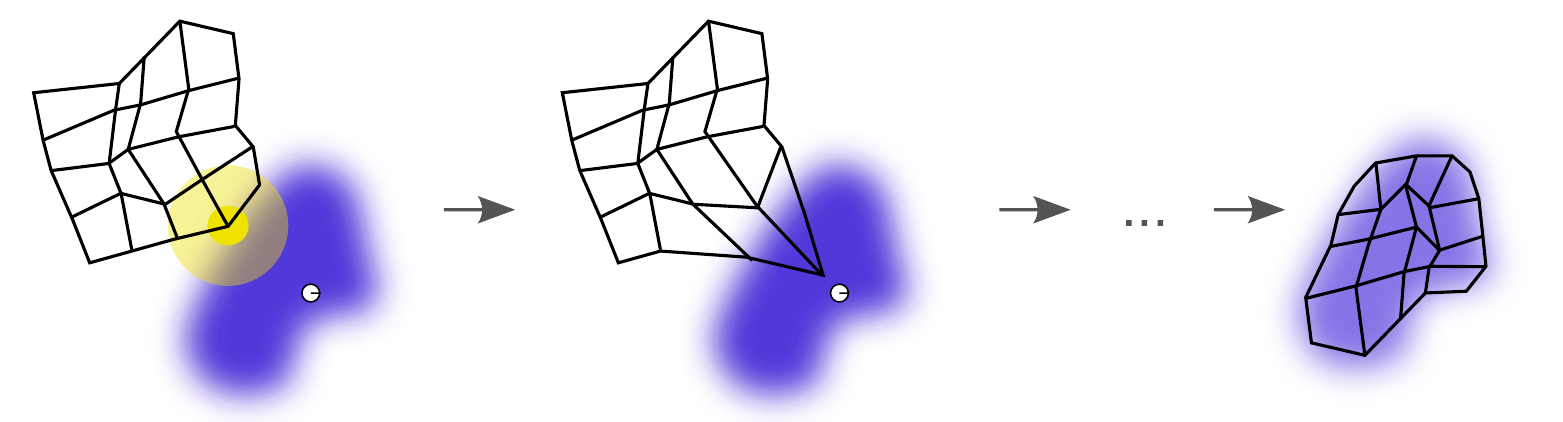}

}

\caption{\label{fig-som-training}Training of a SOM, illustrated on
several steps. Image extracted from
\href{https://en.wikipedia.org/wiki/Self-organizing_map}{Wikipedia}.}

\end{figure}

Figure~\ref{fig-som-training} summarizes and illustrates the training of
a SOM. The blue shape represents the data distribution that we wish to
learn, from a 2D space for easier visualization. Typically, data would
live in higher dimension spaces. Within the data distribution, a white
disc shows the data point that is presented to the SOM at the current
iteration step. SOM neurons, represented by black nodes, and connected
to their neighbors by black edges, are updated towards the current data
point. Among them, the Best Matching Unit, identified by an opaque
yellow disc, is the closest to the current data point, and as such
receives the most important update. The closest neighbors of the BMU,
belonging to the larger yellow transparent disc, are also slightly
updated. Farther neurons are almost not updated. The learned SOM is
represented on the right side of the figure, in which neurons correctly
cover the data distribution.

The update received by a neuron is determined by
Equation~\ref{eq-som-update}, with \(v\) being the index of the neuron,
\(\mathbf{W_v}\) is the prototype vector of neuron \(v\),
\(\mathbf{D_t}\) is the data point presented to the SOM at step \(t\).
\(u\) is the index of the Best Matching Unit, i.e., the neuron that
satisfies
\(u = \argmin_{\forall v} \left\|\mathbf{D_t} - \mathbf{W_v}\right\|\).

\begin{equation}\protect\hypertarget{eq-som-update}{}{
  \mathbf{W_{v}^{t+1}} \gets \mathbf{W_{v}^{t}} + \theta(u, v, t) \alpha(t) \left(\mathbf{D_t} - \mathbf{W_{v}^{t}}\right)
}\label{eq-som-update}\end{equation}

In this equation, \(\theta\) is the neighborhood function, which is
typically a gaussian centered on the BMU (\(u\)), such that the BMU is
the most updated, its closest neighbors are slightly updated, and
farther neurons are not updated. The learning rate \(\alpha\), and the
neighborhood function \(\theta\) both depend on the time step \(t\):
they are often monotonically decreasing, in order to force neurons'
convergence and stability.

One of the numerous extensions of the Self-Organizing Map is the Dynamic
Self-Organizing Map (DSOM) (Rougier and Boniface 2011). The idea behind
DSOMs is that self-organization should offer both stability, when the
input data does not change much, and dynamism, when there is a sudden
change. This stems from neurological inspiration, since the human brain
is able to both stabilize after the early years of development, and
dynamically re-organize itself and adapt when lesions occur.

As we mentioned, the SOM enforces stability through decreasing
parameters (learning rate and neighborhood), however this also prevents
dynamism. Indeed, as the parameters approach \(0\), the vectors' updates
become negligible, and the system does not adapt any more, even when
faced with an abrupt change in the data distribution.

DSOMs propose to replace the time-dependent parameters by a
time-invariant one, named the \emph{elasticity}, which determines the
coupling of neurons. Whereas SOMs and other similar algorithms try to
learn the density of data, DSOMs focus on the structure of the data
space, and the map will not try to place several neurons in a
high-density region. In other words, if a neuron is considered as
sufficiently close to the input data point, the DSOM will not update the
other neurons, assuming that this region of the latent space is already
quite well represented by this neuron. The ``sufficiently close'' is
determined through the elasticity parameter: with high elasticity,
neurons are tightly coupled with each other, whereas lower elasticity
let neurons spread out over the whole latent space.

DSOMs replace the update equation with the following:

\begin{equation}\protect\hypertarget{eq-dsom-update}{}{
  \mathbf{W_i^{t+1}} \leftarrow \alpha\left\|\mathbf{D_t} - \mathbf{W_i^t}\right\|h_{\eta}(i, u, \mathbf{D_t})\left(\mathbf{D_t} - \mathbf{W_i^t}\right)
}\label{eq-dsom-update}\end{equation}

\begin{equation}\protect\hypertarget{eq-dsom-elasticity}{}{
  h_{\eta}(i, u, \mathbf{D_t}) = \exp\left( - \frac{1}{\eta^{2}} \frac{\left\| \pos(i) - \pos(u) \right\|^{2}}{\left\| \mathbf{D_t} - \mathbf{W_u} \right\|^{2}} \right)
}\label{eq-dsom-elasticity}\end{equation}

where \(\alpha\) is the learning rate, \(i\) is the index of the
currently updated neuron, \(\mathbf{D_t}\) is the current data point,
\(u\) is the index of the best matching unit, \(\eta\) is the elasticity
parameter, \(h_{\eta}\) is the neighborhood function, and \(\pos(i)\),
\(\pos(u)\) are respectively the positions of neurons \(i\) and \(u\) in
the grid (not in the latent space). Intuitively, the distance between
\(\pos(i)\) and \(\pos(u)\) is the minimal number of consecutive
neighbors that form a path between \(i\) and \(u\).

\hypertarget{sec-learning-algorithms}{%
\subsection{The learning and decision
algorithms}\label{sec-learning-algorithms}}

We take inspiration from Decentralized Partially-Observable Markovian
Decision Processes (DecPOMDPs) to formally describe our proposed
algorithms. DecPOMDPs are an extension of the well-known Markovian
Decision Process (MDP) that considers multiple agents taking repeated
decisions in multiple states of an environment, by receiving only
partial observations about the current state. In contrast with the
original DecPOMDP as described by Bernstein (Bernstein et al. 2002), we
explicitly define the set of learning agents, and we assume that agents
receive (different) individual rewards, instead of a team reward.

\begin{definition}

A Decentralized Partially-Observable Markovian Decision Process is a
tuple
\(\left\langle \LAgts, \StateSpace, \ActionSpace, \texttt{T}, \ObsSpace, \texttt{O}, \RewardFn, \gamma \right\rangle\),
where:

\begin{itemize}
\tightlist
\item
  \(\LAgts\) is the set of learning agents, of size
  \(n = \length{\LAgts}\).
\item
  \(\StateSpace\) is the state space, i.e., the set of states that the
  environment can possibly be in. States are not directly accessible to
  learning agents.
\item
  \(\ActionSpace_l\) is the set of actions accessible to agent \(l\),
  \(\forall l \in \LAgts\) as all agents take individual actions. We
  consider multi-dimensional and continuous actions, thus we have
  \(\ActionSpace_{l} \subseteq \RR^d\), with \(d\) the number of
  dimensions, which depends on the case of application.
\item
  \(\ActionSpace\) is the action space, i.e., the set of joint-actions
  that can be taken at each time step. A joint-action is the combination
  of all agents' actions, i.e.,
  \(\ActionSpace = \ActionSpace_{l_1} \times \cdots \times \ActionSpace_{l_n}\).
\item
  \(\texttt{T}\) is the transition function, defined as
  \(\texttt{T} : \StateSpace \times \ActionSpace \times \StateSpace \rightarrow [0,1]\).
  In other words, \(\texttt{T}(s' | s, \mathbf{a})\) is the probability
  of obtaining state \(\mathbf{s}'\) after taking the action
  \(\mathbf{a}\) in state \(\mathbf{s}\).
\item
  \(\ObsSpace\) is the observation space, i.e., the set of possible
  observations that agents can receive. An observation is a partial
  information about the current state. Similarly to actions, we define
  \(\ObsSpace_{l}\) as the observation space for learning agent \(l\),
  \(\forall l \in \LAgts\). As well as actions, observations are
  multi-dimensional and continuous, thus we have
  \(\ObsSpace_{l} \subseteq \RR^{g}\), with \(g\) the number of
  dimensions, which depends on the use case.
\item
  \(\texttt{O}\) is the observation probability function, defined as
  \(\texttt{O} : \ObsSpace \times \StateSpace \times \ActionSpace \rightarrow [0,1]\),
  i.e., \(\texttt{O}(\mathbf{o} | s', \mathbf{a})\) is the probability
  of receiving the observations \(\mathbf{o}\) after taking the action
  \(\mathbf{a}\) and arriving in state \(s'\).
\item
  \(\RewardFn\) is the reward function, defined as
  \(\forall l \in \LAgts \quad \RewardFn_{l} : \StateSpace \times \ActionSpace_l \rightarrow \RR\).
  Typically, the reward function itself will be the same for all agents,
  however, agents are rewarded individually, based on their own
  contribution to the environment through their action. In other words,
  \(\RewardFn_{l}(s, \mathbf{a}_{l})\) is the reward that learning agent
  \(l\) receives for taking action \(\mathbf{a}_{l}\) in state \(s\).
\item
  \(\gamma\) is the discount factor, to allow for potentially infinite
  horizon of time steps, with \(\gamma \in [0,1[\).
\end{itemize}

\end{definition}

The RL algorithm must learn a stochastic strategy \(\pi_{l}\), defined
as
\(\pi_{l} : \ObsSpace_{l} \times \ActionSpace_{l} \rightarrow [0,1]\).
In other words, given the observations \(\mathbf{o}_{l}\) received by an
agent \(l\), \(\pi(\mathbf{o}_{l}, \mathbf{a})\) is the probability that
agent \(l\) will take action \(\mathbf{a}\).

We recall that observations and actions are vectors of floating numbers,
the RL algorithm must therefore handle this accordingly. However, it was
mentioned in Section~\ref{sec-learning-qtable} that the Q-Table is not
suitable for continuous data. To solve this, we take inspiration from an
existing work (Smith 2002a, 2002b) and propose to use variants of
Self-Organizing Maps (SOMs) (Kohonen 1990).

We can leverage SOMs to learn to handle the observation and action
spaces: neurons learn the topology of the latent space and create a
discretization. By associating each neuron with a unique index, we are
able to discretize the multi-dimensional data: each data point is
recognized by the neuron with the closest prototype vector, and thus is
represented by a discrete identifier, i.e., the neuron's index.

The proposed algorithms are thus based on two (Dynamic) SOMs, a
State-SOM, and an Action-SOM, which are associated to a Q-Table. To
navigate the Q-Table and access the Q-Values, we use discrete
identifiers obtained from the SOMs. The Q-Table's dimensions thus depend
on the (D)SOMs' number of neurons: the Q-Table has exactly as many rows
as the State-(D)SOM has neurons, and exactly as many columns as the
Action-(D)SOM has neurons, such that each neuron is represented by a row
or column, and reciprocally.

Our algorithms are separated into two distinct parts: the
\emph{decision} process, which chooses an action from received
observations about the environment, and the \emph{learning} process,
which updates the algorithms' data structures, so that the next decision
step will yield a better action. We present in details these two parts
below.

\hypertarget{the-decision-process}{%
\subsubsection{The decision process}\label{the-decision-process}}

\begin{algorithm}
\caption{Decision algorithm}

% Rename the `Require` command to get a sort of `Data` as in algorithm2e
\algrenewcommand\algorithmicrequire{\textbf{Data}}

\label{alg-qsom-decision}
\begin{algorithmic}[1]
\Require $\mathcal{U}$ the neurons in the State-(D)SOM\\
  $\quad \mathbf{U_i}$ the vector associated to neuron $i$ in the State-(D)SOM\\
  $\quad \mathcal{W}$ the neurons in the Action-(D)SOM\\
  $\quad \mathbf{W_j}$ the vector associated to neuron $j$ in the Action-(D)SOM\\
  $\quad \Q(s,a)$ the Q-value of action $a$ in state $s$\\
  $\quad \tau$ the Boltzmann's temperature\\
  $\quad \epsilon$ Noise control parameter\\

\Function{Decision}{Observations $\mathbf{o}$}

  % \Comment{Determine the Best Matching Unit, closest neuron from the State-SOM to the observations}
  \State $s \gets \argmin_{\forall i \in \mathcal{U}} || \mathbf{o} - \mathbf{U_{i}} ||$
  % \Comment{Choose action identifier using Boltzmann probabilities}
  \State Let P be the Boltzmann distribution over the Q-Values. We draw a random variable X from P, and we denote the probability that X equals a given value j : $P(X = j)$.
  \State Draw $j \sim P(X = j) = \frac{\frac{exp(\Q(s,j))}{\tau}}{\sum_{k = 1}^{\length{\mathcal{W}}} \frac{exp(\Q(s,k))}{\tau}}$
  \State Let $\mathbf{W_j}$ be the chosen action's parameters
  % \Comment{Randomly noise the action's parameters to explore}
  \For{$k \in$ all dimensions of $\mathbf{W_j}$}
    % \Comment{The random distribution can be either uniform or normal}
    \State $\mathit{noise} \sim \texttt{random}(\epsilon)$\;
    \State $W'_{j,k} \gets W_{j,k} + \mathit{noise}$\;
  \EndFor
  \State Return action $\mathbf{a} \gets \mathbf{W'_j}$
\EndFunction

\end{algorithmic}
\end{algorithm}

\begin{figure}[htbp]

{\centering \includegraphics[width=1\textwidth,height=\textheight]{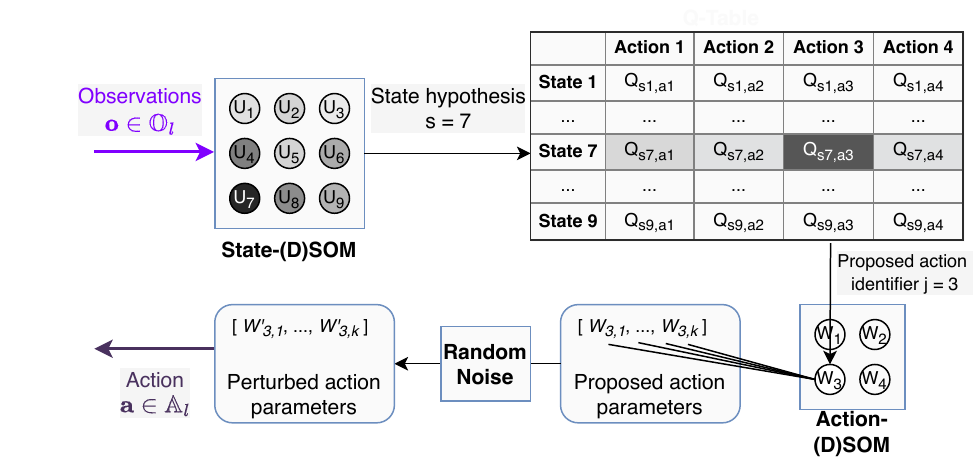}

}

\caption{\label{fig-qsom-decision-schema}Dataflow of the Q-(D)SOM
decision process.}

\end{figure}

We now explain the decision process that allows an agent to choose an
action from received observations, which is described formally in
 Algorithm~\ref{alg-qsom-decision}  and represented in
Figure~\ref{fig-qsom-decision-schema}. First, we need to obtain a
discrete identifier from an observation \(\mathbf{o}\) that is a vector
\(\in \ObsSpace_l \subseteq \RR^{g}\), in order to access the Q-Table.
To do so, we look for the Best Matching Unit (BMU), i.e., the neuron
whose prototype vector is the closest to the observations, from the
State-SOM, which is the SOM tasked with learning the observation space.
The unique index of the BMU is used as the state identifier \(s\) (line
9).

We call this identifier a ``state hypothesis'', and we use it to
navigate the Q-Table and obtain the expected interest of each action,
assuming we have correctly identified the state. Knowing these interests
\(\Q(s,.)\) for all actions, we can assign a probability of taking each
one, using a Boltzmann distribution (line 10). Boltzmann is a well-known
and used method in RL that helps with the exploration-exploitation
dilemma. Indeed, agents should try to maximize their expectancy of
received rewards, which means they should \emph{exploit} high-rewarding
actions, i.e., those with a high interest. However, the true interest of
the action is not known to agents: they have to discover it
incrementally by trying actions into the environment, in various
situations, and memorizing the associated reward. If they only choose
the action with the maximum interest, they risk focusing on few actions,
thus not exploring the others. By not sufficiently exploring, they
maintain the phenomenon, as not explored actions will stay at a low
interest, reducing their probability of being chosen, and so on. Using
Boltzmann mitigates this problem, by giving similar probabilities to
similar interests, and yet, a non-zero probability of being chosen even
for actions with low interests.

The Boltzmann probability of an action \(j\) being selected is computed
based on the action's interest, in the current state, relatively to all
other actions' interests, as follows:

\begin{equation}\protect\hypertarget{eq-boltzmann-probabilities}{}{
  P(X = j) =
\frac{
  \frac{\exp(\Q(s,j))}{\tau}
  }{\sum_{k = 1}^{\length{\mathcal{W}}} \frac{\exp(\Q(s,k))}{\tau}}
}\label{eq-boltzmann-probabilities}\end{equation}

Traditionally, the Boltzmann parameter \(\tau\) should be decreasing
over the time steps, such that the probabilities of high-interest
actions will rise, whereas low-interest actions will converge towards a
probability of \(0\). This mechanism ensures the convergence of the
agents' policy towards the optimal one, by reducing exploration in later
steps, in favour of exploitation. However, and as we have already
mentioned, we chose to disable the convergence mechanisms in our
algorithms, because it prevents, by principle, continuous learning and
adaptation.

We draw an action identifier \(j\) from the list of possible actions,
according to Boltzmann probabilities (line 11). From this discrete
identifier, we get the action's parameters from the Action-SOM, which is
tasked with learning the action space. We retrieve the neuron with
identifier \(j\), and take its prototype vector as the proposed action's
parameters (line 12).

We can note that this is somewhat symmetrical to what is done with the
State-SOM. To learn the State-SOM, we use the data points, i.e., the
observations, that come from the environment; to obtain a discrete
identifier, we take the neurone with the closest prototype. For the
Action-SOM, we start with a discrete identifier, and we take the
prototype of the neuron with this identifier. However, we need to learn
what are those prototype vectors. We do not have data points as for the
State-SOM, since we do not know what is the ``correct'' action in each
situation. In order to learn better actions, we apply an exploration
step after choosing an action: the action's parameters are perturbed by
a random noise (lines 13-16).

In the original work of Smith (2002a), the noise was taken from a
uniform distribution \(\mathcal{U}_{[-\epsilon,+\epsilon]}\), which we
will call the \emph{epsilon} method in our experiments. However, in our
algorithms, we implemented a normal, or \emph{gaussian}, random
distribution \(\mathcal{N}(\mu, \sigma^2)\), where \(\mu\) is the mean,
which we set to \(0\) so that the distribution ranges over both negative
and positive values, \(\sigma^2\) is the variance, and \(\sigma\) is the
standard deviation. \(\epsilon\) and \(\sigma^2\) are the ``noise
control parameter'' for their respective distribution. The advantage
over the uniform distribution is to have a higher probability of a small
noise, thus exploring very close actions, while still allowing for a few
rare but longer ``jumps'' in the action space. These longer jumps may
help to escape local extremas, but should be rare, so as to slowly
converge towards optimal actions most of the time, without overshooting
them. This was not permitted by the uniform distribution, as the
probability is the same for each value in the range
\([-\epsilon,+\epsilon]\).

The noised action's parameters are considered as the chosen action by
the decision process, and the agent executes this action in the
environment (line 17).

\hypertarget{the-learning-process}{%
\subsubsection{The learning process}\label{the-learning-process}}

After all agents executed their action, and the environment simulated
the new state, agents receive a reward signal which indicates to which
degree their action was a ``good one''. From this reward, agents should
improve their behaviour so that their next choice will be better. The
learning process that makes this possible is formally described in
 Algorithm~\ref{alg-qsom-learning} , and we detail it below.

\begin{algorithm}
\caption{Learning algorithm}

% Rename the `Require` command to get a sort of `Data` as in algorithm2e
\algrenewcommand\algorithmicrequire{\textbf{Data}}

\label{alg-qsom-learning}
\begin{algorithmic}[1]
\Require $\mathcal{U}$ the neurons in the State-(D)SOM\\
  $\quad \mathbf{U_u}$ the vector associated to neuron $u$ in the State-(D)SOM\\
  $\quad \mathcal{W}$ the neurons in the Action-(D)SOM\\
  $\quad \mathbf{W_w}$ the vector associated to neuron $w$ in the Action-(D)SOM\\
  $\quad \pos_U(u)$ is the position of neuron $u$ in the State-(D)SOM grid\\
  $\quad \pos_W(w)$ is the position of neuron $w$ in the Action-(D)SOM grid\\
  $\quad \Q(s,a)$ the Q-value of action $a$ in state $s$\\
  $\quad \eta_{U}, \eta_{W}$ elasticity for State- and Action-(D)SOMs\\
  $\quad \alpha_{Q}, \alpha_{U}, \alpha_{W}$ learning rates for Q-Table, State-, Action-(D)SOMs\\
  $\quad \gamma$ the discount factor

\Function{Learning}{Previous observations $\mathbf{o}$, New observations $\mathbf{o}'$, Received reward $r$, State hypothesis $s$, Chosen action identifier $j$, Chosen action parameters $\mathbf{a}$}

  % \Comment{Compute the neighborhood of neurons}
  \For{$u \in \mathcal{U}$}
    \State $\psi_{U}(u) \gets exp\left(-\frac{1}{\eta_{U}^{2}}\frac{\norm{\pos_{U}(u) - \pos_{U}(s)}}{\norm{\mathbf{o} - \mathbf{U_u}}}\right)$
  \EndFor
  \For{$w \in \mathcal{W}$}
    \State $\psi_{W}(w) \gets exp\left(-\frac{1}{\eta_{W}^{2}}\frac{\norm{\pos_{W}(w) - \pos_{W}(j)}}{\norm{\mathbf{a} - \mathbf{W_w}}}\right)$
  \EndFor
  % \Comment{If the action was interesting}
  \If{$r + \gamma \max_{j'} \Q(s',j') \stackrel{?}{>} \Q(s,j)$}
    % \Comment{Update the Action-(D)SOM}
    \For{neuron $w \in \mathcal{W}$}
      \State $\mathbf{W_{w}} \gets \alpha_{W}\norm{\mathbf{a} - \mathbf{W_{w}}}\psi_{W}(w)\left(\mathbf{a} - \mathbf{W_{w}}\right) + \mathbf{W_{w}}$
    \EndFor
  \EndIf
  % \Comment{Update the Q-Table}
  \State $\Q(s,j) \gets \alpha_{Q}\psi_{U}(s)\psi_{W}(j)\left[r + \gamma \max_{j'}\Q(i',j') - \Q(s,j)\right] + \Q(s,j)$
  % \Comment{Update the State-(D)SOM}
  \For{neuron $u \in \mathcal{U}$}
    \State $\mathbf{U_{u}} \gets \alpha_{U}\norm{\mathbf{o} - \mathbf{U_{u}}}\psi_{U}(u)\left(\mathbf{o} - \mathbf{U_{u}}\right) + \mathbf{U_{u}}$
  \EndFor

\EndFunction

\end{algorithmic}
\end{algorithm}

First, we compute the Action-(D)SOM and State-(D)SOM neighborhoods
(lines 11-13 and 14-16). Then, we update the Action-(D)SOM. Remember
that we do not have the ground-truth for actions: we do not know which
parameters yield the best rewards. Moreover, we explored the action
space by randomly noising the proposed action; it is possible that the
perturbed action is actually worse than the learned one. In this case,
we do not want to update the Action-(D)SOM, as this would worsen the
agent's performances. We thus determine whether the perturbed action is
better than the proposed action by comparing the received reward with
the memorized interest of the proposed action, using the following
equation:

\begin{equation}\protect\hypertarget{eq-interesting-proposed-action}{}{
  r + \gamma \max_{j'} \Q(s',j') \stackrel{?}{>} \Q(s,j)
}\label{eq-interesting-proposed-action}\end{equation}

If the perturbed action is deemed better than the proposed one, we
update the Action-(D)SOM towards the perturbed action (lines 17-21). To
do so, we assume that the Best Matching Unit (BMU), i.e., the center of
the neighborhood, is the neuron that was selected at the decision step,
\(j\). We then apply the corresponding update equation,
Equation~\ref{eq-som-update} for a SOM, or Equation~\ref{eq-dsom-update}
for a DSOM, to move the neurons' prototypes towards the perturbed
action.

Secondly, we update the actions' interests, i.e., the Q-Table (line 22).
To do so, we rely on the traditional Bellman's equation. However,
Smith's algorithm introduces a difference in this equation to increase
the learning speed. Indeed, the State- and Action-(D)SOMs offer
additional knowledge about the states and actions: as they are discrete
identifiers mapping to continuous vectors in a latent space, we can
define a notion of \emph{similarity} between states (resp. actions) by
measuring the distance between the states' vectors (resp. actions'
vectors). Similar states and actions will most likely have a similar
interest, and thus each Q-Value is updated at each time step, instead of
only the current state-action pair, by taking into account the
neighborhoods of the State- and Action-(D)SOMs (computed on lines 11-13
and 14-16). Equation~\ref{eq-neighborhood-bellman} shows the resulting
formula:

\begin{equation}\protect\hypertarget{eq-neighborhood-bellman}{}{
  \Q_{t+1}(s,j) \leftarrow \alpha \psi_{U}(s) \psi_{W}(j) \left[r + \gamma \max_{j'} \Q_{t}(s',j') \right] + (1 - \alpha)\Q_{t}(s,j)
}\label{eq-neighborhood-bellman}\end{equation}

where \(s\) was the state hypothesis at step \(t\), \(j\) was the chosen
action identifier, \(r\) is the received reward, \(s'\) is the state
hypothesis at step \(t+1\) (from the new observations). \(\psi_U(s)\)
and \(\psi_W(j)\) represent, respectively, the neighborhood of the
State- and Action-(D)SOMs, centered on the state \(s\) and the chosen
action identifier \(j\). Intuitively, the equation takes into account
the interest of arriving in this new state, based on the maximum
interest of actions available in the new state. This means that an
action could yield a medium reward by itself, but still be very
interesting because it allows to take actions with higher interests. On
the contrary, an action with a high reward, but leading to a state with
only catastrophic actions would have a low interest.

Finally, we learn the State-SOM, which is a very simple step (lines
23-25). Indeed, we have already mentioned that we know data points,
i.e., observations, that have been sampled from the distribution of
states by the environment. Therefore, we simply update the neurons'
prototypes towards the received observation at the previous step.
Prototype vectors are updated based on both their own distance to the
data point, within the latent space, and the distance between their
neuron and the best matching unit, within the 2D grid neighborhood
(using the neighborhood computed on lines 11-13). This ensures that the
State-SOM learns to represent states which appear in the environment.

\begin{remark}

In the presented algorithm, the neighborhood and update formulas
correspond to a DSOM. When using the QSOM algorithm, these formulas must
be replaced by their SOM equivalents. The general structure of the
algorithm, i.e., the steps and the order in which they are taken, stays
the same.

\end{remark}

\begin{remark}

Compared to Smith's algorithm, our extensions differ in the following
aspects:

\begin{itemize}
\tightlist
\item
  DSOMs can be used in addition to SOMs.
\item
  Hyperparameters are not annealed, i.e., they are constant throughout
  the simulation, so that agents can continuously learn instead of
  slowly converging.
\item
  Actions are chosen through a Boltzmann distribution of probabilities
  based on their interests, instead of using the \(\epsilon\)-greedy
  method.
\item
  The random noise to explore the actions' space is drawn from a
  Gaussian distribution instead of a uniform one.
\item
  The neighborhood functions of the State- and Action-(D)SOMs is a
  gaussian instead of a linear one.
\item
  The number of dimensions of the actions' space in the following
  experiments is greater (6) than in Smith's original experiments (2).
  This particularly prompted the need to explore other ways to randomly
  noise actions, e.g., the gaussian distribution. Note that some other
  methods have been tried, such as applying a noise on a single
  dimension each step, or randomly determining for each dimension
  whether it should be noised at each step; they are not disclosed in
  the results as they performed slightly below the gaussian method.
  Searching for better hyperparameters could yield better results for
  these methods.
\end{itemize}

\end{remark}

\hypertarget{sec-experiments}{%
\section{Experiments and results}\label{sec-experiments}}

In order to validate our proposed algorithms, we ran some experiments on
a Smart Grid use-case..

First, let us apply the algorithms and formal model on this specific
use-case. The observation space, \(\ObsSpace\), is composed of the
information that agents receive: the time (hour), the available energy,
their personal battery storage, \ldots{} The full list of observations
was defined in Section @ref(positioning-smartgrid-observations). These
values range from 0 to 1, and we have 11 such values, thus we define
\(\ObsSpace_l = [0,1]^{11}\).

Similarly, actions are defined by multiple parameters: consume energy
from grid, consume from battery, sell, \ldots{} These actions were
presented in Section @ref(positioning-smartgrid-actions). To simplify
the learning of actions, we constrain these parameters to the \([0,1]\)
range; they are scaled to the true agent's action range outside the
learning and decision processes. For example, let us imagine an agent
with an action range of \(6,000\), and an action parameter, as outputted
by the decision process, of \(0.5\), the scaled action parameter will be
\(0.5 \times 6,000 = 3,000\). We have 6 actions parameters, and thus
define \(\ActionSpace_l = [0,1]^{6}\).

In the sequel, we present the reward functions that we implemented to
test our algorithms, as well as the experiments' scenarii. Finally, we
quickly describe the 2 algorithms that we chose as baselines:
\emph{DDPG} and \emph{MADDPG}.

\hypertarget{sec-smartgrid}{%
\subsection{The Smart-Grid use-case}\label{sec-smartgrid}}

\begin{figure}[htbp]

{\centering \includegraphics{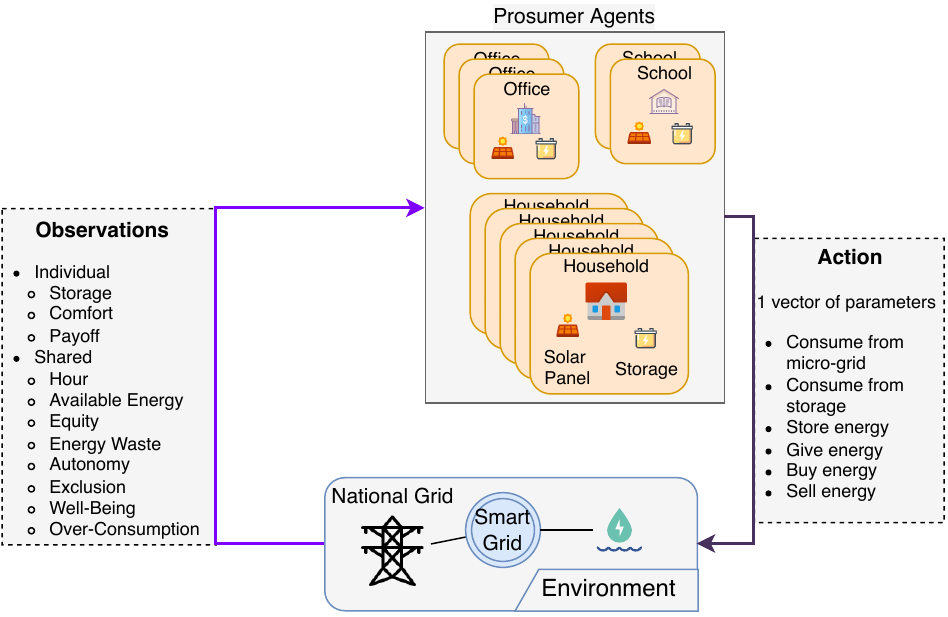}

}

\caption{\label{fig-smartgrid}Illustration of the Smart Grid use-case.
Multiple learning agents receive observations, and must decide to take
actions, in order to exchange energy.}

\end{figure}

We use, to evaluate the QSOM and QDSOM algorithms, a Smart-Grid use case
in which multiple producer-consumer (\emph{prosumer}) agents learn to
consume energy to satisfy their needs. The use-case is represented in
Figure~\ref{fig-smartgrid}.

Learning agents receive \emph{observations} \(\in \RR^{11}\) that
describe the current state of the environment: they consist of
\emph{shared} observations \(\in \RR^{8}\) that are shared among all
agents, such as the current hour or amount of available energy, and
\emph{local} observations \(\in \RR^{3}\) that are individual to each
agent, and not accessible to others, such as the agent's personal
battery. Splitting between shared and local observations helps
preserving the privacy of agents, by not sharing personal data. From
these observations, agents must take \emph{actions}, represented by
vectors of parameters \(\in \RR^{6}\), which govern the amounts of
energy to exchange: how much to consume from the smart grid, how much to
buy, etc.

In practice, these observations and actions are interpolated to the
\([0,1]\) domain, so as to facilitate the learning algorithms, and
especially the Self-Organizing Maps. Indeed, a dimension with a higher
range than another would have a greater importance and would risk
biasing the learning.

The simulated Smart Grid is connected to a national grid, which allows
agents to buy and sell energy, although from more pollutant sources; it
is also connected to an hydropower plant, which is considered to be
local to the Smart Grid. This power plant generates the energy that is
available to all agents at each time step of the simulation. Agents
additionally produce a small quantity (e.g., from solar panels), which
is kept in their personal battery. They may share this energy with other
agents when necessary (e.g., to increase equity), consume it directly,
or sell it to the national grid for some (monetary) profit.

Different \emph{profiles} of prosumer agents are present in the grid,
each representing a specific kind of building: a (small) Household, an
(medium) Office, or a (large) School. Buildings' profiles determine
several characteristics, such as the \emph{needs} that these buildings
have, i.e., how much energy they would like to consume at each hour.
These needs are taken from a public dataset of energy consumption in the
United States (Ong and Clark 2014). Profiles also impact the range of
action parameters: larger buildings may consume more energy than the
smaller ones. In practice, the range was determined to be slightly
higher than the maximum need over all hours, so that the buildings can
decide to consume as much as they need (yet, during the simulation, this
might be a bad idea due to the environment's state!). Similarly, the
battery capacity depends on the profile, with larger buildings having
access to higher capacities.

Agents make decisions based on the rewards they receive, which drive
them towards the respect of one or several ethical considerations. The
reward functions are described in the next section, and concern some
considerations that are classical for smart grid: consuming energy to
satisfy their needs and increase their comfort, ensuring the equity of
comforts among agents, avoiding to over-consume.

\hypertarget{sec-learning-experiments-rewards}{%
\subsection{Reward functions}\label{sec-learning-experiments-rewards}}

We implemented multiple reward functions that each focus on different
ethical stakes. Most of them are based on the principle of Difference
Reward (Yliniemi and Tumer 2014) to facilitate the Credit Assignment.
Additionally, two functions focus on multiple objectives, but with a
rather naïve approach to scalarize, and another two focus on adaptation,
i.e., the agents' capacity to adapt their behaviour to changing mores,
by making the reward function artificially change at a fixed point in
time.

We give an intuitive definition and a mathematical formula for each of
these reward functions below.

\begin{description}
\tightlist
\item[Equity]
Determine the agent's contribution to the society's equity, by comparing
the current equity with the equity if the agent did not act. The agent's
goal is thus to maximize the society's equity.

\[\RewardFn_{eq}(agent) = (1 - \texttt{Hoover}(Comforts)) - (1 - \texttt{Hoover}(Comforts \setminus \{agent\}))\]
\item[Over-Consumption]
Determine the agent's contribution to over-consumption, by comparing the
current over-consumed amount of energy, with the amount that would have
been over-consumed if the agent did not act. The agent's goal is thus to
minimize society's over-consumption.

\[\RewardFn_{oc}(agent) = 1 - \frac{\mathit{OC}}{\sum_{\forall a} (\mathit{Consumed}_{a} + \mathit{Stored}_{a})} - \frac{\mathit{OC} - (\mathit{Consumed}_{agent} + \mathit{Stored}_{agent})}{\sum_{\forall a \neq agent} (\mathit{Consumed}_{a} + \mathit{Stored}_{a})}\]
\item[Comfort]
Simply return the agent's comfort, so that agents aim to maximize their
comfort. This intuitively does not seem like an ethical stake, however
it can be linked to Schwartz' ``hedonistic'' value, and therefore is an
ethical stake, focused on the individual aspect. We will mainly use this
reward function in combination with others that focus on the societal
aspect, to demonstrate the algorithms' capacity to learn opposed moral
values.

\[\RewardFn_{comfort}(agent) = Comforts_{agent}\]
\item[Multi-Objective Sum]
A first and simple reward function that combines multiple objectives,
namely limitation of over-consumption and comfort. The goal of agents is
thus to both minimize the society's over-consumption while maximizing
their own comfort. This may be a difficult task, because the simulation
is designed so that there is a scarcity of energy most of the time, and
agents will most likely over-consume if they all try to maximize their
comfort. On the contrary, reducing the over-consumption means they need
to diminish their comfort. There is thus a trade-off to be achieved
between over-consumption and comfort.

\[\RewardFn_{mos}(agent) = 0.8 \times \RewardFn_{oc}(agent) + 0.2 \times \RewardFn_{comfort}(agent)\]
\item[Multi-Objective Product]
A second, but also simple, multi-objective reward functions. Instead of
using a weighted sum, we multiply the reward together. This function is
more punitive than the sum, as a low reward cannot be ``compensated''.
For example, let us consider a vector of reward components
\([0.1, 0.9]\). Using the weighted sum, the result depends on the
weights: if the first component has a low coefficient, then the result
may actually be high. On contrary, the product will return
\(0.1 \times 0.9 = 0.09\), i.e., a very low reward. Any low component
will penalize the final result.

\[\RewardFn_{mop}(agent) = \RewardFn_{oc}(agent) \times \RewardFn_{comfort}(agent)\]
\item[Adaptability1]
A reward function that simulates a change in its definition after 2000
time steps, as if society's ethical mores had changed. During the first
2000 steps, it behaves similarly as the Over-Consumption reward
function, whereas for later steps it returns the mean of
Over-Consumption and Equity rewards.

\[\RewardFn_{ada1}(agent) = \begin{cases}
  \RewardFn_{oc}(agent) & \text{if } t < 2000 \\
  \frac{\RewardFn_{oc}(agent) + \RewardFn_{eq}(agent)}{2} & \text{else}
\end{cases}\]
\item[Adaptability2]
Similar to Adaptability1, this function simulates a change in its
definition. We increase the difficulty by making 2 changes, one after
2000 time steps, and another after 6000 time steps, and by considering a
combination of 3 rewards after the second change.

\[\RewardFn_{ada2}(agent) = \begin{cases}
  \RewardFn_{oc}(agent) & \text{if } t < 2000 \\
  \frac{\RewardFn_{oc}(agent) + \RewardFn_{eq}(agent)}{2} & \text{else if } t < 6000 \\
  \frac{\RewardFn_{oc}(agent) + \RewardFn_{eq}(agent) + \RewardFn_{comfort}(agent)}{3} & \text{else}
\end{cases}\]
\end{description}

As we can see, the various reward functions have different aims. Some
simple functions, such as \emph{equity}, \emph{overconsumption}, or
\emph{comfort}, serve as a baseline and building blocks for other
functions. Nevertheless, they may be easy to optimize: for example, by
consuming absolutely nothing, the \emph{overconsumption} function can be
satisifed. On the contrary, the \emph{comfort} function can be satisfied
by consuming the maximum amount of energy, such that the comfort is
guaranteed to be close to \(1\). The two \emph{multi-objective}
functions thus try to force agents to learn several stakes at the same
time, especially if they are contradictory, such as
\emph{overconsumption} and \emph{comfort}. The agent thus cannot learn a
``trivial'' behaviour and must find the optimal behaviour that manages
to satisfy both as much as possible. Finally, the \emph{adaptability}
functions go a step further and evaluate agents' ability to adapt when
the considerations change.

\hypertarget{sec-learning-experiments-scenarii}{%
\subsection{Scenarii}\label{sec-learning-experiments-scenarii}}

In order to improve the richness of our experiments, we designed several
scenarii. These scenarii are defined by two variables: the agents'
consumption profile, and the environment's size, i.e., number of agents.

\begin{figure}[htbp]

{\centering \includegraphics[width=0.9\textwidth,height=\textheight]{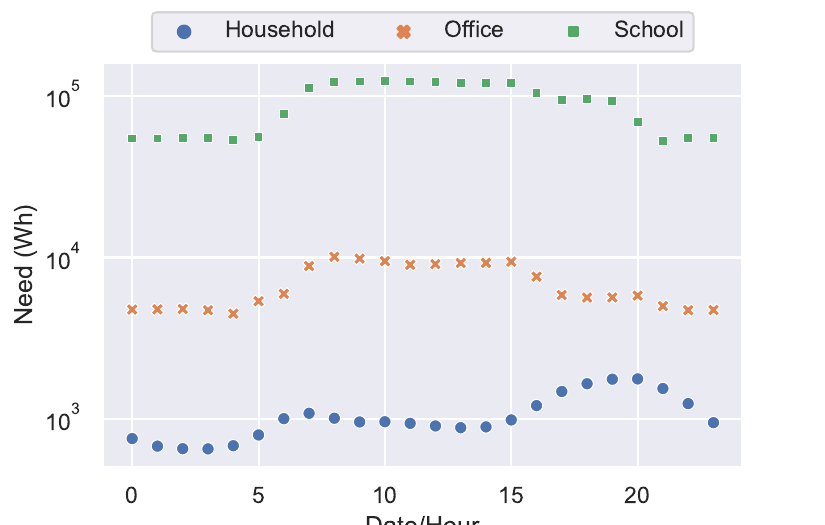}

}

\caption{\label{fig-learning-scenarii-plot-needs}The agents' needs for
every hour of the day in the \emph{daily} profile.}

\end{figure}

The prosumer (learning) agents are instantiated with a profile,
determining their battery capacity, their action range, and their needs,
i.e., the quantity of energy they want to consume at each hour. These
needs are extracted from real consumption profiles; we propose two
different versions, the \emph{daily} and the \emph{annual} profiles. In
the \emph{daily} version, needs are averaged over every day of the year,
thus yielding a need for each hour of a day: this is illustrated in
Figure~\ref{fig-learning-scenarii-plot-needs}. This is a simplified
version, averaging the seasonal differences; its advantages are a
reduced size, thus decreasing the required computational resources, a
simpler learning, and an easier visualization for humans. On the other
hand, the \emph{annual} version is more complete, contains seasonal
differences, which improve the environment's richness and force agents
to adapt to important changes.

The second property is the environment size. We wanted to test our
algorithms with different sets of agents, to ensure the scalability of
the approach, in the sense that agents are able to learn a correct
behaviour and adapt to many other agents in the same environment. This
may be difficult as the number of agents increases, since there will
most certainly be more conflicts. We propose a first, \emph{small}
environment, containing \(20\) Households agents, \(5\) Office agents,
and \(1\) School agent. The second environment, \emph{medium}, contains
roughly 4 times more agents than in the \emph{small} case: \(80\)
Household agents, \(19\) Office agents, and \(1\) School.

\hypertarget{learning-experiments-baselines}{%
\subsubsection{DDPG and MADDPG
baselines}\label{learning-experiments-baselines}}

In order to prove our algorithms' advantages, we chose to compare them
to the well-known DDPG (Lillicrap et al. 2015) and its multi-agent
extension, MADDPG (Lowe et al. 2017).

DDPG (\emph{Deep Deterministic Policy Gradient}) is one of the
algorithms that extended the success of Deep Reinforcement Learning to
continuous domains (Lillicrap et al. 2015). It follows the quite popular
Actor-Critic architecture, which uses two different Neural Networks: one
for the Actor, i.e., to decide which action to perform at each time
step, and another for the Critic, i.e., to evaluate whether an action is
interesting. We chose it as a baseline since it focuses on problems with
similar characteristics, e.g., continuous domains, and is a popular
baseline in the community.

MADDPG (\emph{Multi-Agent Deep Deterministic Policy Gradient}), extends
the idea of DDPG to the multi-agent setting (Lowe et al. 2017), by
relying on the \emph{Centralized Training - Decentralized Execution}
idea. It is one of the most used methods to improve multi-agent
learning, by sharing data among agents during the learning phase. This
helps agents make a model of other agents and adapt to their respective
behaviours. However, during execution, sharing data in the same manner
is often impracticable or undesirable, as it would impair privacy and
require some sort of communication between agents; thus, data is not
shared any more at this point (Decentralized Execution). As such,
Centralized Training - Decentralized Execution makes a distinction
between training and execution, and is thus inadequate for continuous
learning, and constant adaptation to changes. On the other hand, if we
were to make agents continuously learn with centralized data sharing,
even in the execution phase, we would impair privacy of users that are
represented or impacted by the agents. These reasons are why we chose
not to use this setting for our own algorithms QSOM and QDSOM. While we
do not use centralized training, we want to compare them to an algorithm
that uses it, such as MADDPG, in order to determine whether there would
be a performance gain, and what would be the trade-off between
performance and privacy. In MADDPG, the Centralized Training is simply
done by using a centralized Critic network, which receives observations,
actions, and rewards from all agents, and evaluates all agents' actions.
The Actor networks, however, are still individualized: each agent has
its own network, which the other agents cannot access. During the
training phase, the Critic network is updated thanks to the globally
shared data, whereas Actor networks are updated through local data and
the global Critic. Once the learning is done, the networks are frozen:
the Critic does not require receiving global data any more, and the
Actors do not rely on the Critic any more. Only the decision part, i.e.,
which action should we do, is kept, by using the trained Actor network
as-is.

\hypertarget{learning-results}{%
\subsection{Results}\label{learning-results}}

Several sets of experiments were performed:

\begin{itemize}
\item
  First, numerous experiments were launched to search for the best
  hyperparameters of each algorithm, to ensure a fair comparison later.
  Each set of hyperparameters was run \(10\) times to obtain average
  results, and a better statistical significance. In order to limit the
  number of runs and thus the computational resources required, we
  decided to focus on the \emph{adaptability2} reward for these
  experiments. This function is difficult enough so that the algorithms
  will not reach almost 100\% immediately, which would make the
  hyperparameter search quite useless, and is one of the 2 that interest
  us the most, along with \emph{adaptability1}, so it makes sense that
  our algorithms are optimized for this one. The \emph{annual}
  consumption profile was used to increase the richness, but the
  environment size, i.e., number of agents, was set to \emph{small} in
  order to once again reduce the computational power and time.
\item
  Then, the 4 algorithms, configured with their best hyperparameters,
  were compared on multiple settings: both \emph{annual} and
  \emph{daily} consumption profiles, both \emph{small} and \emph{medium}
  sizes of environment, and all the reward functions. This resulted in
  \(2 \times 2 \times 7\) scenarii, which we ran \(10\) times for each
  of the \(4\) algorithms.
\end{itemize}

In the following results, we define a run's \emph{score} as the average
of the global rewards per step. The global reward corresponds to the
reward, without focusing on a specific agent. For example, the
\emph{equity} reward compares the Hoover index of the whole environment
to a hypothetical environment without the agent. The global reward, in
this case, is simply the Hoover index of the entire environment. This
represents, intuitively, how the society of agents performed, globally.
Taking the average is one of the simplest methods to get a single score
for a given run, which allows comparing runs easily.

\hypertarget{searching-for-hyperparameters}{%
\subsubsection{Searching for
hyperparameters}\label{searching-for-hyperparameters}}

Table~\ref{tbl-learning1-best-hyperparameters-qsom},
Table~\ref{tbl-learning1-best-hyperparameters-qdsom},
Table~\ref{tbl-learning1-best-hyperparameters-ddpg}, and
Table~\ref{tbl-learning1-best-hyperparameters-maddpg} summarize the best
hyperparameters that have been found for each algorithm, based on the
average runs' score obtained when using these parameters.

\hypertarget{tbl-learning1-best-hyperparameters-qsom}{}
\begin{table}[H]

\providecommand{\docline}[3]{\noalign{\global\setlength{\arrayrulewidth}{#1}}\arrayrulecolor[HTML]{#2}\cline{#3}}

\setlength{\tabcolsep}{0pt}

\renewcommand*{\arraystretch}{1.5}

\begin{longtable}[c]{|p{1.84in}|p{0.84in}|p{2.96in}}

\caption{\label{tbl-learning1-best-hyperparameters-qsom}Best hyperparameters on 10 runs for the \emph{QSOM} algorithm, using the
\emph{annual small} scenario and \emph{adaptability2} reward function. } \\

\hhline{>{\arrayrulecolor[HTML]{666666}\global\arrayrulewidth=2pt}->{\arrayrulecolor[HTML]{666666}\global\arrayrulewidth=2pt}->{\arrayrulecolor[HTML]{666666}\global\arrayrulewidth=2pt}-}

\multicolumn{1}{!{\color[HTML]{000000}\vrule width 0pt}>{\raggedright}p{\dimexpr 1.84in+0\tabcolsep+0\arrayrulewidth}}{\textcolor[HTML]{000000}{\fontsize{11}{11}\selectfont{Parameter}}} & \multicolumn{1}{!{\color[HTML]{000000}\vrule width 0pt}>{\raggedright}p{\dimexpr 0.84in+0\tabcolsep+0\arrayrulewidth}}{\textcolor[HTML]{000000}{\fontsize{11}{11}\selectfont{Value}}} & \multicolumn{1}{!{\color[HTML]{000000}\vrule width 0pt}>{\raggedright}p{\dimexpr 2.96in+0\tabcolsep+0\arrayrulewidth}!{\color[HTML]{000000}\vrule width 0pt}}{\textcolor[HTML]{000000}{\fontsize{11}{11}\selectfont{Description}}} \\

\hhline{>{\arrayrulecolor[HTML]{666666}\global\arrayrulewidth=2pt}->{\arrayrulecolor[HTML]{666666}\global\arrayrulewidth=2pt}->{\arrayrulecolor[HTML]{666666}\global\arrayrulewidth=2pt}-}\endhead

\multicolumn{1}{!{\color[HTML]{000000}\vrule width 0pt}>{\raggedright}p{\dimexpr 1.84in+0\tabcolsep+0\arrayrulewidth}}{\textcolor[HTML]{000000}{\fontsize{10}{10}\selectfont{State-SOM\ shape}}} & \multicolumn{1}{!{\color[HTML]{000000}\vrule width 0pt}>{\raggedright}p{\dimexpr 0.84in+0\tabcolsep+0\arrayrulewidth}}{\textcolor[HTML]{000000}{\fontsize{10}{10}\selectfont{12x12}}} & \multicolumn{1}{!{\color[HTML]{000000}\vrule width 0pt}>{\raggedright}p{\dimexpr 2.96in+0\tabcolsep+0\arrayrulewidth}!{\color[HTML]{000000}\vrule width 0pt}}{\textcolor[HTML]{000000}{\fontsize{10}{10}\selectfont{Shape\ of\ the\ neurons'\ grid}}} \\

\multicolumn{1}{!{\color[HTML]{000000}\vrule width 0pt}>{\raggedright}p{\dimexpr 1.84in+0\tabcolsep+0\arrayrulewidth}}{\textcolor[HTML]{000000}{\fontsize{10}{10}\selectfont{State-SOM\ learning\ rate}}} & \multicolumn{1}{!{\color[HTML]{000000}\vrule width 0pt}>{\raggedright}p{\dimexpr 0.84in+0\tabcolsep+0\arrayrulewidth}}{\textcolor[HTML]{000000}{\fontsize{10}{10}\selectfont{0.5}}} & \multicolumn{1}{!{\color[HTML]{000000}\vrule width 0pt}>{\raggedright}p{\dimexpr 2.96in+0\tabcolsep+0\arrayrulewidth}!{\color[HTML]{000000}\vrule width 0pt}}{\textcolor[HTML]{000000}{\fontsize{10}{10}\selectfont{Update\ speed\ of\ State-SOM\ neurons}}} \\

\multicolumn{1}{!{\color[HTML]{000000}\vrule width 0pt}>{\raggedright}p{\dimexpr 1.84in+0\tabcolsep+0\arrayrulewidth}}{\textcolor[HTML]{000000}{\fontsize{10}{10}\selectfont{Action-SOM\ shape}}} & \multicolumn{1}{!{\color[HTML]{000000}\vrule width 0pt}>{\raggedright}p{\dimexpr 0.84in+0\tabcolsep+0\arrayrulewidth}}{\textcolor[HTML]{000000}{\fontsize{10}{10}\selectfont{3x3}}} & \multicolumn{1}{!{\color[HTML]{000000}\vrule width 0pt}>{\raggedright}p{\dimexpr 2.96in+0\tabcolsep+0\arrayrulewidth}!{\color[HTML]{000000}\vrule width 0pt}}{\textcolor[HTML]{000000}{\fontsize{10}{10}\selectfont{Shape\ of\ the\ neurons'\ grid}}} \\

\multicolumn{1}{!{\color[HTML]{000000}\vrule width 0pt}>{\raggedright}p{\dimexpr 1.84in+0\tabcolsep+0\arrayrulewidth}}{\textcolor[HTML]{000000}{\fontsize{10}{10}\selectfont{Action-SOM\ learning\ rate}}} & \multicolumn{1}{!{\color[HTML]{000000}\vrule width 0pt}>{\raggedright}p{\dimexpr 0.84in+0\tabcolsep+0\arrayrulewidth}}{\textcolor[HTML]{000000}{\fontsize{10}{10}\selectfont{0.2}}} & \multicolumn{1}{!{\color[HTML]{000000}\vrule width 0pt}>{\raggedright}p{\dimexpr 2.96in+0\tabcolsep+0\arrayrulewidth}!{\color[HTML]{000000}\vrule width 0pt}}{\textcolor[HTML]{000000}{\fontsize{10}{10}\selectfont{Update\ speed\ of\ Action-SOM\ neurons}}} \\

\multicolumn{1}{!{\color[HTML]{000000}\vrule width 0pt}>{\raggedright}p{\dimexpr 1.84in+0\tabcolsep+0\arrayrulewidth}}{\textcolor[HTML]{000000}{\fontsize{10}{10}\selectfont{Q\ Learning\ rate}}} & \multicolumn{1}{!{\color[HTML]{000000}\vrule width 0pt}>{\raggedright}p{\dimexpr 0.84in+0\tabcolsep+0\arrayrulewidth}}{\textcolor[HTML]{000000}{\fontsize{10}{10}\selectfont{0.6}}} & \multicolumn{1}{!{\color[HTML]{000000}\vrule width 0pt}>{\raggedright}p{\dimexpr 2.96in+0\tabcolsep+0\arrayrulewidth}!{\color[HTML]{000000}\vrule width 0pt}}{\textcolor[HTML]{000000}{\fontsize{10}{10}\selectfont{Update\ speed\ of\ Q-Values}}} \\

\multicolumn{1}{!{\color[HTML]{000000}\vrule width 0pt}>{\raggedright}p{\dimexpr 1.84in+0\tabcolsep+0\arrayrulewidth}}{\textcolor[HTML]{000000}{\fontsize{10}{10}\selectfont{Discount\ rate}}} & \multicolumn{1}{!{\color[HTML]{000000}\vrule width 0pt}>{\raggedright}p{\dimexpr 0.84in+0\tabcolsep+0\arrayrulewidth}}{\textcolor[HTML]{000000}{\fontsize{10}{10}\selectfont{0.9}}} & \multicolumn{1}{!{\color[HTML]{000000}\vrule width 0pt}>{\raggedright}p{\dimexpr 2.96in+0\tabcolsep+0\arrayrulewidth}!{\color[HTML]{000000}\vrule width 0pt}}{\textcolor[HTML]{000000}{\fontsize{10}{10}\selectfont{Controls\ the\ horizon\ of\ rewards}}} \\

\multicolumn{1}{!{\color[HTML]{000000}\vrule width 0pt}>{\raggedright}p{\dimexpr 1.84in+0\tabcolsep+0\arrayrulewidth}}{\textcolor[HTML]{000000}{\fontsize{10}{10}\selectfont{Action\ perturbation}}} & \multicolumn{1}{!{\color[HTML]{000000}\vrule width 0pt}>{\raggedright}p{\dimexpr 0.84in+0\tabcolsep+0\arrayrulewidth}}{\textcolor[HTML]{000000}{\fontsize{10}{10}\selectfont{gaussian}}} & \multicolumn{1}{!{\color[HTML]{000000}\vrule width 0pt}>{\raggedright}p{\dimexpr 2.96in+0\tabcolsep+0\arrayrulewidth}!{\color[HTML]{000000}\vrule width 0pt}}{\textcolor[HTML]{000000}{\fontsize{10}{10}\selectfont{Method\ to\ randomly\ explore\ actions}}} \\

\multicolumn{1}{!{\color[HTML]{000000}\vrule width 0pt}>{\raggedright}p{\dimexpr 1.84in+0\tabcolsep+0\arrayrulewidth}}{\textcolor[HTML]{000000}{\fontsize{10}{10}\selectfont{Action\ noise}}} & \multicolumn{1}{!{\color[HTML]{000000}\vrule width 0pt}>{\raggedright}p{\dimexpr 0.84in+0\tabcolsep+0\arrayrulewidth}}{\textcolor[HTML]{000000}{\fontsize{10}{10}\selectfont{0.06}}} & \multicolumn{1}{!{\color[HTML]{000000}\vrule width 0pt}>{\raggedright}p{\dimexpr 2.96in+0\tabcolsep+0\arrayrulewidth}!{\color[HTML]{000000}\vrule width 0pt}}{\textcolor[HTML]{000000}{\fontsize{10}{10}\selectfont{Parameter\ for\ the\ random\ noise\ distribution}}} \\

\multicolumn{1}{!{\color[HTML]{000000}\vrule width 0pt}>{\raggedright}p{\dimexpr 1.84in+0\tabcolsep+0\arrayrulewidth}}{\textcolor[HTML]{000000}{\fontsize{10}{10}\selectfont{Boltzmann\ temperature}}} & \multicolumn{1}{!{\color[HTML]{000000}\vrule width 0pt}>{\raggedright}p{\dimexpr 0.84in+0\tabcolsep+0\arrayrulewidth}}{\textcolor[HTML]{000000}{\fontsize{10}{10}\selectfont{0.4}}} & \multicolumn{1}{!{\color[HTML]{000000}\vrule width 0pt}>{\raggedright}p{\dimexpr 2.96in+0\tabcolsep+0\arrayrulewidth}!{\color[HTML]{000000}\vrule width 0pt}}{\textcolor[HTML]{000000}{\fontsize{10}{10}\selectfont{Controls\ the\ exploration-exploitation}}} \\

\hhline{>{\arrayrulecolor[HTML]{666666}\global\arrayrulewidth=2pt}->{\arrayrulecolor[HTML]{666666}\global\arrayrulewidth=2pt}->{\arrayrulecolor[HTML]{666666}\global\arrayrulewidth=2pt}-}

\end{longtable}

\end{table}

\hypertarget{tbl-learning1-best-hyperparameters-qdsom}{}
\begin{table}[H]

\providecommand{\docline}[3]{\noalign{\global\setlength{\arrayrulewidth}{#1}}\arrayrulecolor[HTML]{#2}\cline{#3}}

\setlength{\tabcolsep}{0pt}

\renewcommand*{\arraystretch}{1.5}

\begin{longtable}[c]{|p{1.94in}|p{0.84in}|p{2.96in}}

\caption{\label{tbl-learning1-best-hyperparameters-qdsom}Best hyperparameters on 10 runs for the \emph{QDSOM} algorithm, using
the \emph{annual small} scenario and \emph{adaptability2} reward
function. } \\

\hhline{>{\arrayrulecolor[HTML]{666666}\global\arrayrulewidth=2pt}->{\arrayrulecolor[HTML]{666666}\global\arrayrulewidth=2pt}->{\arrayrulecolor[HTML]{666666}\global\arrayrulewidth=2pt}-}

\multicolumn{1}{!{\color[HTML]{000000}\vrule width 0pt}>{\raggedright}p{\dimexpr 1.94in+0\tabcolsep+0\arrayrulewidth}}{\textcolor[HTML]{000000}{\fontsize{11}{11}\selectfont{Parameter}}} & \multicolumn{1}{!{\color[HTML]{000000}\vrule width 0pt}>{\raggedright}p{\dimexpr 0.84in+0\tabcolsep+0\arrayrulewidth}}{\textcolor[HTML]{000000}{\fontsize{11}{11}\selectfont{Value}}} & \multicolumn{1}{!{\color[HTML]{000000}\vrule width 0pt}>{\raggedright}p{\dimexpr 2.96in+0\tabcolsep+0\arrayrulewidth}!{\color[HTML]{000000}\vrule width 0pt}}{\textcolor[HTML]{000000}{\fontsize{11}{11}\selectfont{Description}}} \\

\hhline{>{\arrayrulecolor[HTML]{666666}\global\arrayrulewidth=2pt}->{\arrayrulecolor[HTML]{666666}\global\arrayrulewidth=2pt}->{\arrayrulecolor[HTML]{666666}\global\arrayrulewidth=2pt}-}\endhead

\multicolumn{1}{!{\color[HTML]{000000}\vrule width 0pt}>{\raggedright}p{\dimexpr 1.94in+0\tabcolsep+0\arrayrulewidth}}{\textcolor[HTML]{000000}{\fontsize{10}{10}\selectfont{State-DSOM\ shape}}} & \multicolumn{1}{!{\color[HTML]{000000}\vrule width 0pt}>{\raggedright}p{\dimexpr 0.84in+0\tabcolsep+0\arrayrulewidth}}{\textcolor[HTML]{000000}{\fontsize{10}{10}\selectfont{12x12}}} & \multicolumn{1}{!{\color[HTML]{000000}\vrule width 0pt}>{\raggedright}p{\dimexpr 2.96in+0\tabcolsep+0\arrayrulewidth}!{\color[HTML]{000000}\vrule width 0pt}}{\textcolor[HTML]{000000}{\fontsize{10}{10}\selectfont{Shape\ of\ the\ neurons'\ grid}}} \\

\multicolumn{1}{!{\color[HTML]{000000}\vrule width 0pt}>{\raggedright}p{\dimexpr 1.94in+0\tabcolsep+0\arrayrulewidth}}{\textcolor[HTML]{000000}{\fontsize{10}{10}\selectfont{State-DSOM\ learning\ rate}}} & \multicolumn{1}{!{\color[HTML]{000000}\vrule width 0pt}>{\raggedright}p{\dimexpr 0.84in+0\tabcolsep+0\arrayrulewidth}}{\textcolor[HTML]{000000}{\fontsize{10}{10}\selectfont{0.8}}} & \multicolumn{1}{!{\color[HTML]{000000}\vrule width 0pt}>{\raggedright}p{\dimexpr 2.96in+0\tabcolsep+0\arrayrulewidth}!{\color[HTML]{000000}\vrule width 0pt}}{\textcolor[HTML]{000000}{\fontsize{10}{10}\selectfont{Update\ speed\ of\ State-DSOM\ neurons}}} \\

\multicolumn{1}{!{\color[HTML]{000000}\vrule width 0pt}>{\raggedright}p{\dimexpr 1.94in+0\tabcolsep+0\arrayrulewidth}}{\textcolor[HTML]{000000}{\fontsize{10}{10}\selectfont{State-DSOM\ elasticity}}} & \multicolumn{1}{!{\color[HTML]{000000}\vrule width 0pt}>{\raggedright}p{\dimexpr 0.84in+0\tabcolsep+0\arrayrulewidth}}{\textcolor[HTML]{000000}{\fontsize{10}{10}\selectfont{1}}} & \multicolumn{1}{!{\color[HTML]{000000}\vrule width 0pt}>{\raggedright}p{\dimexpr 2.96in+0\tabcolsep+0\arrayrulewidth}!{\color[HTML]{000000}\vrule width 0pt}}{\textcolor[HTML]{000000}{\fontsize{10}{10}\selectfont{Coupling\ between\ State-DSOM\ neurons}}} \\

\multicolumn{1}{!{\color[HTML]{000000}\vrule width 0pt}>{\raggedright}p{\dimexpr 1.94in+0\tabcolsep+0\arrayrulewidth}}{\textcolor[HTML]{000000}{\fontsize{10}{10}\selectfont{Action-DSOM\ shape}}} & \multicolumn{1}{!{\color[HTML]{000000}\vrule width 0pt}>{\raggedright}p{\dimexpr 0.84in+0\tabcolsep+0\arrayrulewidth}}{\textcolor[HTML]{000000}{\fontsize{10}{10}\selectfont{3x3}}} & \multicolumn{1}{!{\color[HTML]{000000}\vrule width 0pt}>{\raggedright}p{\dimexpr 2.96in+0\tabcolsep+0\arrayrulewidth}!{\color[HTML]{000000}\vrule width 0pt}}{\textcolor[HTML]{000000}{\fontsize{10}{10}\selectfont{Shape\ of\ the\ neurons'\ grid}}} \\

\multicolumn{1}{!{\color[HTML]{000000}\vrule width 0pt}>{\raggedright}p{\dimexpr 1.94in+0\tabcolsep+0\arrayrulewidth}}{\textcolor[HTML]{000000}{\fontsize{10}{10}\selectfont{Action-DSOM\ learning\ rate}}} & \multicolumn{1}{!{\color[HTML]{000000}\vrule width 0pt}>{\raggedright}p{\dimexpr 0.84in+0\tabcolsep+0\arrayrulewidth}}{\textcolor[HTML]{000000}{\fontsize{10}{10}\selectfont{0.7}}} & \multicolumn{1}{!{\color[HTML]{000000}\vrule width 0pt}>{\raggedright}p{\dimexpr 2.96in+0\tabcolsep+0\arrayrulewidth}!{\color[HTML]{000000}\vrule width 0pt}}{\textcolor[HTML]{000000}{\fontsize{10}{10}\selectfont{Update\ speed\ of\ Action-DSOM\ neurons}}} \\

\multicolumn{1}{!{\color[HTML]{000000}\vrule width 0pt}>{\raggedright}p{\dimexpr 1.94in+0\tabcolsep+0\arrayrulewidth}}{\textcolor[HTML]{000000}{\fontsize{10}{10}\selectfont{Action-DSOM\ elasticity}}} & \multicolumn{1}{!{\color[HTML]{000000}\vrule width 0pt}>{\raggedright}p{\dimexpr 0.84in+0\tabcolsep+0\arrayrulewidth}}{\textcolor[HTML]{000000}{\fontsize{10}{10}\selectfont{1}}} & \multicolumn{1}{!{\color[HTML]{000000}\vrule width 0pt}>{\raggedright}p{\dimexpr 2.96in+0\tabcolsep+0\arrayrulewidth}!{\color[HTML]{000000}\vrule width 0pt}}{\textcolor[HTML]{000000}{\fontsize{10}{10}\selectfont{Coupling\ between\ Action-DSOM\ neurons}}} \\

\multicolumn{1}{!{\color[HTML]{000000}\vrule width 0pt}>{\raggedright}p{\dimexpr 1.94in+0\tabcolsep+0\arrayrulewidth}}{\textcolor[HTML]{000000}{\fontsize{10}{10}\selectfont{Q\ Learning\ rate}}} & \multicolumn{1}{!{\color[HTML]{000000}\vrule width 0pt}>{\raggedright}p{\dimexpr 0.84in+0\tabcolsep+0\arrayrulewidth}}{\textcolor[HTML]{000000}{\fontsize{10}{10}\selectfont{0.8}}} & \multicolumn{1}{!{\color[HTML]{000000}\vrule width 0pt}>{\raggedright}p{\dimexpr 2.96in+0\tabcolsep+0\arrayrulewidth}!{\color[HTML]{000000}\vrule width 0pt}}{\textcolor[HTML]{000000}{\fontsize{10}{10}\selectfont{Update\ speed\ of\ Q-Values}}} \\

\multicolumn{1}{!{\color[HTML]{000000}\vrule width 0pt}>{\raggedright}p{\dimexpr 1.94in+0\tabcolsep+0\arrayrulewidth}}{\textcolor[HTML]{000000}{\fontsize{10}{10}\selectfont{Discount\ rate}}} & \multicolumn{1}{!{\color[HTML]{000000}\vrule width 0pt}>{\raggedright}p{\dimexpr 0.84in+0\tabcolsep+0\arrayrulewidth}}{\textcolor[HTML]{000000}{\fontsize{10}{10}\selectfont{0.95}}} & \multicolumn{1}{!{\color[HTML]{000000}\vrule width 0pt}>{\raggedright}p{\dimexpr 2.96in+0\tabcolsep+0\arrayrulewidth}!{\color[HTML]{000000}\vrule width 0pt}}{\textcolor[HTML]{000000}{\fontsize{10}{10}\selectfont{Controls\ the\ horizon\ of\ rewards}}} \\

\multicolumn{1}{!{\color[HTML]{000000}\vrule width 0pt}>{\raggedright}p{\dimexpr 1.94in+0\tabcolsep+0\arrayrulewidth}}{\textcolor[HTML]{000000}{\fontsize{10}{10}\selectfont{Action\ perturbation}}} & \multicolumn{1}{!{\color[HTML]{000000}\vrule width 0pt}>{\raggedright}p{\dimexpr 0.84in+0\tabcolsep+0\arrayrulewidth}}{\textcolor[HTML]{000000}{\fontsize{10}{10}\selectfont{gaussian}}} & \multicolumn{1}{!{\color[HTML]{000000}\vrule width 0pt}>{\raggedright}p{\dimexpr 2.96in+0\tabcolsep+0\arrayrulewidth}!{\color[HTML]{000000}\vrule width 0pt}}{\textcolor[HTML]{000000}{\fontsize{10}{10}\selectfont{Method\ to\ randomly\ explore\ actions}}} \\

\multicolumn{1}{!{\color[HTML]{000000}\vrule width 0pt}>{\raggedright}p{\dimexpr 1.94in+0\tabcolsep+0\arrayrulewidth}}{\textcolor[HTML]{000000}{\fontsize{10}{10}\selectfont{Action\ noise}}} & \multicolumn{1}{!{\color[HTML]{000000}\vrule width 0pt}>{\raggedright}p{\dimexpr 0.84in+0\tabcolsep+0\arrayrulewidth}}{\textcolor[HTML]{000000}{\fontsize{10}{10}\selectfont{0.09}}} & \multicolumn{1}{!{\color[HTML]{000000}\vrule width 0pt}>{\raggedright}p{\dimexpr 2.96in+0\tabcolsep+0\arrayrulewidth}!{\color[HTML]{000000}\vrule width 0pt}}{\textcolor[HTML]{000000}{\fontsize{10}{10}\selectfont{Parameter\ for\ the\ random\ noise\ distribution}}} \\

\multicolumn{1}{!{\color[HTML]{000000}\vrule width 0pt}>{\raggedright}p{\dimexpr 1.94in+0\tabcolsep+0\arrayrulewidth}}{\textcolor[HTML]{000000}{\fontsize{10}{10}\selectfont{Boltzmann\ temperature}}} & \multicolumn{1}{!{\color[HTML]{000000}\vrule width 0pt}>{\raggedright}p{\dimexpr 0.84in+0\tabcolsep+0\arrayrulewidth}}{\textcolor[HTML]{000000}{\fontsize{10}{10}\selectfont{0.6}}} & \multicolumn{1}{!{\color[HTML]{000000}\vrule width 0pt}>{\raggedright}p{\dimexpr 2.96in+0\tabcolsep+0\arrayrulewidth}!{\color[HTML]{000000}\vrule width 0pt}}{\textcolor[HTML]{000000}{\fontsize{10}{10}\selectfont{Controls\ the\ exploration-exploitation}}} \\

\hhline{>{\arrayrulecolor[HTML]{666666}\global\arrayrulewidth=2pt}->{\arrayrulecolor[HTML]{666666}\global\arrayrulewidth=2pt}->{\arrayrulecolor[HTML]{666666}\global\arrayrulewidth=2pt}-}

\end{longtable}

\end{table}

\hypertarget{tbl-learning1-best-hyperparameters-ddpg}{}
\begin{table}[H]

\providecommand{\docline}[3]{\noalign{\global\setlength{\arrayrulewidth}{#1}}\arrayrulecolor[HTML]{#2}\cline{#3}}

\setlength{\tabcolsep}{0pt}

\renewcommand*{\arraystretch}{1.5}

\begin{longtable}[c]{|p{1.45in}|p{0.84in}|p{3.40in}}

\caption{\label{tbl-learning1-best-hyperparameters-ddpg}Best hyperparameters on 10 runs for the \emph{DDPG} algorithm, using the
\emph{annual small} scenario and \emph{adaptability2} reward function. } \\

\hhline{>{\arrayrulecolor[HTML]{666666}\global\arrayrulewidth=2pt}->{\arrayrulecolor[HTML]{666666}\global\arrayrulewidth=2pt}->{\arrayrulecolor[HTML]{666666}\global\arrayrulewidth=2pt}-}

\multicolumn{1}{!{\color[HTML]{000000}\vrule width 0pt}>{\raggedright}p{\dimexpr 1.45in+0\tabcolsep+0\arrayrulewidth}}{\textcolor[HTML]{000000}{\fontsize{11}{11}\selectfont{Parameter}}} & \multicolumn{1}{!{\color[HTML]{000000}\vrule width 0pt}>{\raggedright}p{\dimexpr 0.84in+0\tabcolsep+0\arrayrulewidth}}{\textcolor[HTML]{000000}{\fontsize{11}{11}\selectfont{Value}}} & \multicolumn{1}{!{\color[HTML]{000000}\vrule width 0pt}>{\raggedright}p{\dimexpr 3.4in+0\tabcolsep+0\arrayrulewidth}!{\color[HTML]{000000}\vrule width 0pt}}{\textcolor[HTML]{000000}{\fontsize{11}{11}\selectfont{Description}}} \\

\hhline{>{\arrayrulecolor[HTML]{666666}\global\arrayrulewidth=2pt}->{\arrayrulecolor[HTML]{666666}\global\arrayrulewidth=2pt}->{\arrayrulecolor[HTML]{666666}\global\arrayrulewidth=2pt}-}\endhead

\multicolumn{1}{!{\color[HTML]{000000}\vrule width 0pt}>{\raggedright}p{\dimexpr 1.45in+0\tabcolsep+0\arrayrulewidth}}{\textcolor[HTML]{000000}{\fontsize{10}{10}\selectfont{Batch\ size}}} & \multicolumn{1}{!{\color[HTML]{000000}\vrule width 0pt}>{\raggedright}p{\dimexpr 0.84in+0\tabcolsep+0\arrayrulewidth}}{\textcolor[HTML]{000000}{\fontsize{10}{10}\selectfont{256}}} & \multicolumn{1}{!{\color[HTML]{000000}\vrule width 0pt}>{\raggedright}p{\dimexpr 3.4in+0\tabcolsep+0\arrayrulewidth}!{\color[HTML]{000000}\vrule width 0pt}}{\textcolor[HTML]{000000}{\fontsize{10}{10}\selectfont{Number\ of\ samples\ to\ use\ for\ training\ at\ each\ step}}} \\

\multicolumn{1}{!{\color[HTML]{000000}\vrule width 0pt}>{\raggedright}p{\dimexpr 1.45in+0\tabcolsep+0\arrayrulewidth}}{\textcolor[HTML]{000000}{\fontsize{10}{10}\selectfont{Learning\ rate}}} & \multicolumn{1}{!{\color[HTML]{000000}\vrule width 0pt}>{\raggedright}p{\dimexpr 0.84in+0\tabcolsep+0\arrayrulewidth}}{\textcolor[HTML]{000000}{\fontsize{10}{10}\selectfont{5e-04}}} & \multicolumn{1}{!{\color[HTML]{000000}\vrule width 0pt}>{\raggedright}p{\dimexpr 3.4in+0\tabcolsep+0\arrayrulewidth}!{\color[HTML]{000000}\vrule width 0pt}}{\textcolor[HTML]{000000}{\fontsize{10}{10}\selectfont{Update\ speed\ of\ neural\ networks}}} \\

\multicolumn{1}{!{\color[HTML]{000000}\vrule width 0pt}>{\raggedright}p{\dimexpr 1.45in+0\tabcolsep+0\arrayrulewidth}}{\textcolor[HTML]{000000}{\fontsize{10}{10}\selectfont{Discount\ rate}}} & \multicolumn{1}{!{\color[HTML]{000000}\vrule width 0pt}>{\raggedright}p{\dimexpr 0.84in+0\tabcolsep+0\arrayrulewidth}}{\textcolor[HTML]{000000}{\fontsize{10}{10}\selectfont{0.99}}} & \multicolumn{1}{!{\color[HTML]{000000}\vrule width 0pt}>{\raggedright}p{\dimexpr 3.4in+0\tabcolsep+0\arrayrulewidth}!{\color[HTML]{000000}\vrule width 0pt}}{\textcolor[HTML]{000000}{\fontsize{10}{10}\selectfont{Controls\ the\ horizon\ of\ rewards}}} \\

\multicolumn{1}{!{\color[HTML]{000000}\vrule width 0pt}>{\raggedright}p{\dimexpr 1.45in+0\tabcolsep+0\arrayrulewidth}}{\textcolor[HTML]{000000}{\fontsize{10}{10}\selectfont{Tau}}} & \multicolumn{1}{!{\color[HTML]{000000}\vrule width 0pt}>{\raggedright}p{\dimexpr 0.84in+0\tabcolsep+0\arrayrulewidth}}{\textcolor[HTML]{000000}{\fontsize{10}{10}\selectfont{5e-04}}} & \multicolumn{1}{!{\color[HTML]{000000}\vrule width 0pt}>{\raggedright}p{\dimexpr 3.4in+0\tabcolsep+0\arrayrulewidth}!{\color[HTML]{000000}\vrule width 0pt}}{\textcolor[HTML]{000000}{\fontsize{10}{10}\selectfont{Target\ network\ update\ rate}}} \\

\multicolumn{1}{!{\color[HTML]{000000}\vrule width 0pt}>{\raggedright}p{\dimexpr 1.45in+0\tabcolsep+0\arrayrulewidth}}{\textcolor[HTML]{000000}{\fontsize{10}{10}\selectfont{Action\ perturbation}}} & \multicolumn{1}{!{\color[HTML]{000000}\vrule width 0pt}>{\raggedright}p{\dimexpr 0.84in+0\tabcolsep+0\arrayrulewidth}}{\textcolor[HTML]{000000}{\fontsize{10}{10}\selectfont{gaussian}}} & \multicolumn{1}{!{\color[HTML]{000000}\vrule width 0pt}>{\raggedright}p{\dimexpr 3.4in+0\tabcolsep+0\arrayrulewidth}!{\color[HTML]{000000}\vrule width 0pt}}{\textcolor[HTML]{000000}{\fontsize{10}{10}\selectfont{Method\ to\ randomly\ explore\ actions}}} \\

\multicolumn{1}{!{\color[HTML]{000000}\vrule width 0pt}>{\raggedright}p{\dimexpr 1.45in+0\tabcolsep+0\arrayrulewidth}}{\textcolor[HTML]{000000}{\fontsize{10}{10}\selectfont{Action\ noise}}} & \multicolumn{1}{!{\color[HTML]{000000}\vrule width 0pt}>{\raggedright}p{\dimexpr 0.84in+0\tabcolsep+0\arrayrulewidth}}{\textcolor[HTML]{000000}{\fontsize{10}{10}\selectfont{0.11}}} & \multicolumn{1}{!{\color[HTML]{000000}\vrule width 0pt}>{\raggedright}p{\dimexpr 3.4in+0\tabcolsep+0\arrayrulewidth}!{\color[HTML]{000000}\vrule width 0pt}}{\textcolor[HTML]{000000}{\fontsize{10}{10}\selectfont{Parameter\ for\ the\ random\ noise\ distribution}}} \\

\hhline{>{\arrayrulecolor[HTML]{666666}\global\arrayrulewidth=2pt}->{\arrayrulecolor[HTML]{666666}\global\arrayrulewidth=2pt}->{\arrayrulecolor[HTML]{666666}\global\arrayrulewidth=2pt}-}

\end{longtable}

\end{table}

\hypertarget{tbl-learning1-best-hyperparameters-maddpg}{}
\begin{table}[H]

\providecommand{\docline}[3]{\noalign{\global\setlength{\arrayrulewidth}{#1}}\arrayrulecolor[HTML]{#2}\cline{#3}}

\setlength{\tabcolsep}{0pt}

\renewcommand*{\arraystretch}{1.5}

\begin{longtable}[c]{|p{1.42in}|p{0.68in}|p{3.40in}}

\caption{\label{tbl-learning1-best-hyperparameters-maddpg}Best hyperparameters on 10 runs for the \emph{MADDPG} algorithm, using
the \emph{annual small} scenario and \emph{adaptability2} reward
function. } \\

\hhline{>{\arrayrulecolor[HTML]{666666}\global\arrayrulewidth=2pt}->{\arrayrulecolor[HTML]{666666}\global\arrayrulewidth=2pt}->{\arrayrulecolor[HTML]{666666}\global\arrayrulewidth=2pt}-}

\multicolumn{1}{!{\color[HTML]{000000}\vrule width 0pt}>{\raggedright}p{\dimexpr 1.42in+0\tabcolsep+0\arrayrulewidth}}{\textcolor[HTML]{000000}{\fontsize{11}{11}\selectfont{Parameter}}} & \multicolumn{1}{!{\color[HTML]{000000}\vrule width 0pt}>{\raggedright}p{\dimexpr 0.68in+0\tabcolsep+0\arrayrulewidth}}{\textcolor[HTML]{000000}{\fontsize{11}{11}\selectfont{Value}}} & \multicolumn{1}{!{\color[HTML]{000000}\vrule width 0pt}>{\raggedright}p{\dimexpr 3.4in+0\tabcolsep+0\arrayrulewidth}!{\color[HTML]{000000}\vrule width 0pt}}{\textcolor[HTML]{000000}{\fontsize{11}{11}\selectfont{Description}}} \\

\hhline{>{\arrayrulecolor[HTML]{666666}\global\arrayrulewidth=2pt}->{\arrayrulecolor[HTML]{666666}\global\arrayrulewidth=2pt}->{\arrayrulecolor[HTML]{666666}\global\arrayrulewidth=2pt}-}\endhead

\multicolumn{1}{!{\color[HTML]{000000}\vrule width 0pt}>{\raggedright}p{\dimexpr 1.42in+0\tabcolsep+0\arrayrulewidth}}{\textcolor[HTML]{000000}{\fontsize{10}{10}\selectfont{Batch\ size}}} & \multicolumn{1}{!{\color[HTML]{000000}\vrule width 0pt}>{\raggedright}p{\dimexpr 0.68in+0\tabcolsep+0\arrayrulewidth}}{\textcolor[HTML]{000000}{\fontsize{10}{10}\selectfont{128}}} & \multicolumn{1}{!{\color[HTML]{000000}\vrule width 0pt}>{\raggedright}p{\dimexpr 3.4in+0\tabcolsep+0\arrayrulewidth}!{\color[HTML]{000000}\vrule width 0pt}}{\textcolor[HTML]{000000}{\fontsize{10}{10}\selectfont{Number\ of\ samples\ to\ use\ for\ training\ at\ each\ step}}} \\

\multicolumn{1}{!{\color[HTML]{000000}\vrule width 0pt}>{\raggedright}p{\dimexpr 1.42in+0\tabcolsep+0\arrayrulewidth}}{\textcolor[HTML]{000000}{\fontsize{10}{10}\selectfont{Buffer\ size}}} & \multicolumn{1}{!{\color[HTML]{000000}\vrule width 0pt}>{\raggedright}p{\dimexpr 0.68in+0\tabcolsep+0\arrayrulewidth}}{\textcolor[HTML]{000000}{\fontsize{10}{10}\selectfont{50000}}} & \multicolumn{1}{!{\color[HTML]{000000}\vrule width 0pt}>{\raggedright}p{\dimexpr 3.4in+0\tabcolsep+0\arrayrulewidth}!{\color[HTML]{000000}\vrule width 0pt}}{\textcolor[HTML]{000000}{\fontsize{10}{10}\selectfont{Size\ of\ the\ replay\ memory.}}} \\

\multicolumn{1}{!{\color[HTML]{000000}\vrule width 0pt}>{\raggedright}p{\dimexpr 1.42in+0\tabcolsep+0\arrayrulewidth}}{\textcolor[HTML]{000000}{\fontsize{10}{10}\selectfont{Actor\ learning\ rate}}} & \multicolumn{1}{!{\color[HTML]{000000}\vrule width 0pt}>{\raggedright}p{\dimexpr 0.68in+0\tabcolsep+0\arrayrulewidth}}{\textcolor[HTML]{000000}{\fontsize{10}{10}\selectfont{0.01}}} & \multicolumn{1}{!{\color[HTML]{000000}\vrule width 0pt}>{\raggedright}p{\dimexpr 3.4in+0\tabcolsep+0\arrayrulewidth}!{\color[HTML]{000000}\vrule width 0pt}}{\textcolor[HTML]{000000}{\fontsize{10}{10}\selectfont{Update\ speed\ of\ the\ Actor\ network}}} \\

\multicolumn{1}{!{\color[HTML]{000000}\vrule width 0pt}>{\raggedright}p{\dimexpr 1.42in+0\tabcolsep+0\arrayrulewidth}}{\textcolor[HTML]{000000}{\fontsize{10}{10}\selectfont{Critic\ learning\ rate}}} & \multicolumn{1}{!{\color[HTML]{000000}\vrule width 0pt}>{\raggedright}p{\dimexpr 0.68in+0\tabcolsep+0\arrayrulewidth}}{\textcolor[HTML]{000000}{\fontsize{10}{10}\selectfont{0.001}}} & \multicolumn{1}{!{\color[HTML]{000000}\vrule width 0pt}>{\raggedright}p{\dimexpr 3.4in+0\tabcolsep+0\arrayrulewidth}!{\color[HTML]{000000}\vrule width 0pt}}{\textcolor[HTML]{000000}{\fontsize{10}{10}\selectfont{Update\ speed\ of\ the\ Critic\ network}}} \\

\multicolumn{1}{!{\color[HTML]{000000}\vrule width 0pt}>{\raggedright}p{\dimexpr 1.42in+0\tabcolsep+0\arrayrulewidth}}{\textcolor[HTML]{000000}{\fontsize{10}{10}\selectfont{Discount\ rate}}} & \multicolumn{1}{!{\color[HTML]{000000}\vrule width 0pt}>{\raggedright}p{\dimexpr 0.68in+0\tabcolsep+0\arrayrulewidth}}{\textcolor[HTML]{000000}{\fontsize{10}{10}\selectfont{0.95}}} & \multicolumn{1}{!{\color[HTML]{000000}\vrule width 0pt}>{\raggedright}p{\dimexpr 3.4in+0\tabcolsep+0\arrayrulewidth}!{\color[HTML]{000000}\vrule width 0pt}}{\textcolor[HTML]{000000}{\fontsize{10}{10}\selectfont{Controls\ the\ horizon\ of\ rewards}}} \\

\multicolumn{1}{!{\color[HTML]{000000}\vrule width 0pt}>{\raggedright}p{\dimexpr 1.42in+0\tabcolsep+0\arrayrulewidth}}{\textcolor[HTML]{000000}{\fontsize{10}{10}\selectfont{Tau}}} & \multicolumn{1}{!{\color[HTML]{000000}\vrule width 0pt}>{\raggedright}p{\dimexpr 0.68in+0\tabcolsep+0\arrayrulewidth}}{\textcolor[HTML]{000000}{\fontsize{10}{10}\selectfont{0.001}}} & \multicolumn{1}{!{\color[HTML]{000000}\vrule width 0pt}>{\raggedright}p{\dimexpr 3.4in+0\tabcolsep+0\arrayrulewidth}!{\color[HTML]{000000}\vrule width 0pt}}{\textcolor[HTML]{000000}{\fontsize{10}{10}\selectfont{Target\ network\ update\ rate}}} \\

\multicolumn{1}{!{\color[HTML]{000000}\vrule width 0pt}>{\raggedright}p{\dimexpr 1.42in+0\tabcolsep+0\arrayrulewidth}}{\textcolor[HTML]{000000}{\fontsize{10}{10}\selectfont{Noise}}} & \multicolumn{1}{!{\color[HTML]{000000}\vrule width 0pt}>{\raggedright}p{\dimexpr 0.68in+0\tabcolsep+0\arrayrulewidth}}{\textcolor[HTML]{000000}{\fontsize{10}{10}\selectfont{0.02}}} & \multicolumn{1}{!{\color[HTML]{000000}\vrule width 0pt}>{\raggedright}p{\dimexpr 3.4in+0\tabcolsep+0\arrayrulewidth}!{\color[HTML]{000000}\vrule width 0pt}}{\textcolor[HTML]{000000}{\fontsize{10}{10}\selectfont{Controls\ a\ gaussian\ noise\ to\ explore\ actions}}} \\

\multicolumn{1}{!{\color[HTML]{000000}\vrule width 0pt}>{\raggedright}p{\dimexpr 1.42in+0\tabcolsep+0\arrayrulewidth}}{\textcolor[HTML]{000000}{\fontsize{10}{10}\selectfont{Epsilon}}} & \multicolumn{1}{!{\color[HTML]{000000}\vrule width 0pt}>{\raggedright}p{\dimexpr 0.68in+0\tabcolsep+0\arrayrulewidth}}{\textcolor[HTML]{000000}{\fontsize{10}{10}\selectfont{0.05}}} & \multicolumn{1}{!{\color[HTML]{000000}\vrule width 0pt}>{\raggedright}p{\dimexpr 3.4in+0\tabcolsep+0\arrayrulewidth}!{\color[HTML]{000000}\vrule width 0pt}}{\textcolor[HTML]{000000}{\fontsize{10}{10}\selectfont{Controls\ the\ exploration-exploitation}}} \\

\hhline{>{\arrayrulecolor[HTML]{666666}\global\arrayrulewidth=2pt}->{\arrayrulecolor[HTML]{666666}\global\arrayrulewidth=2pt}->{\arrayrulecolor[HTML]{666666}\global\arrayrulewidth=2pt}-}

\end{longtable}

\end{table}

\hypertarget{comparing-algorithms}{%
\subsubsection{Comparing algorithms}\label{comparing-algorithms}}

\begin{figure}[!ht]

{\centering \includegraphics[width=1\textwidth,height=\textheight]{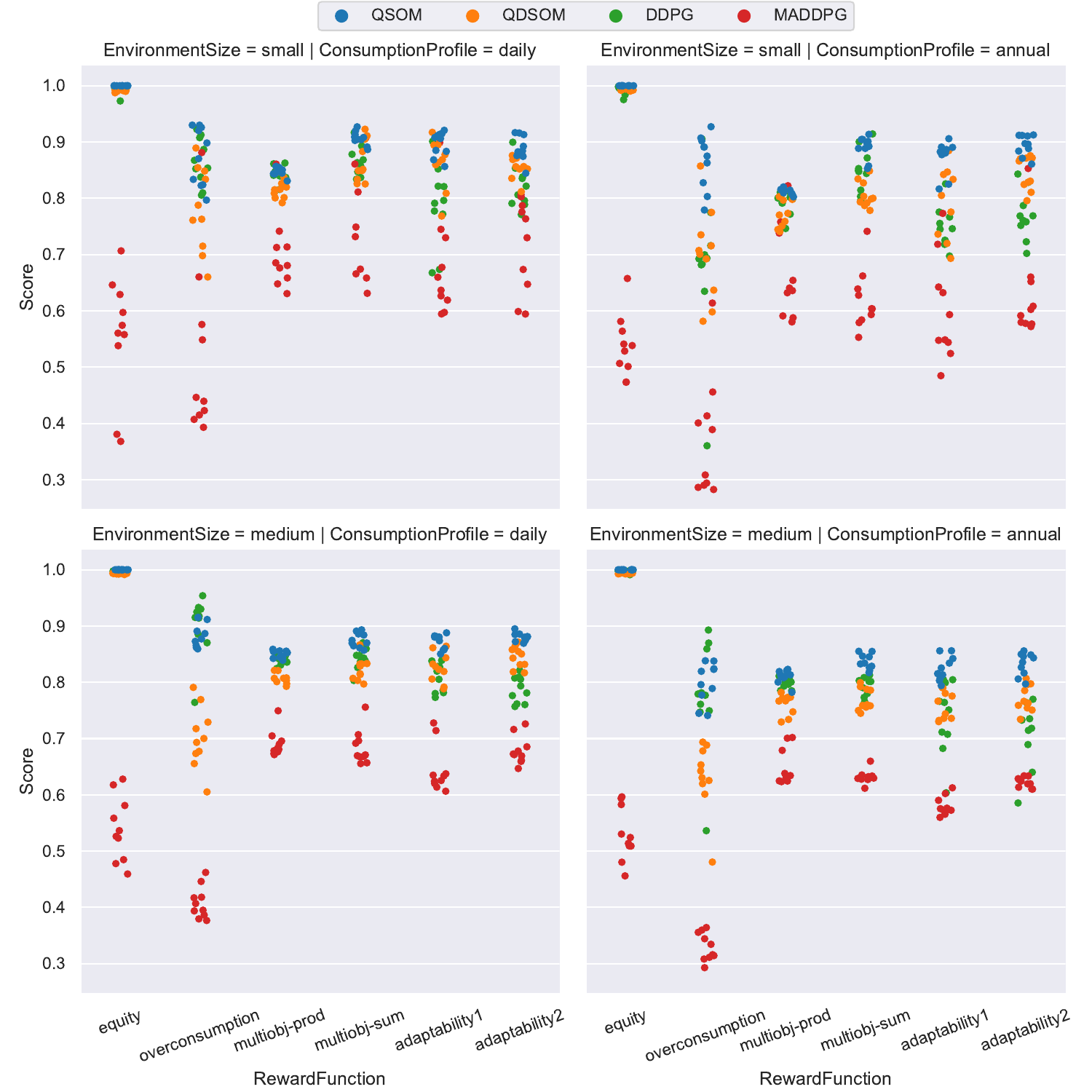}

}

\caption{\label{fig-learning1-results-compare}Distribution of scores per
learning algorithm, on every scenario, for 10 runs with each reward
function.}

\end{figure}

\hypertarget{tbl-learning1-results-compare-table}{}
\begin{table}[H]

\providecommand{\docline}[3]{\noalign{\global\setlength{\arrayrulewidth}{#1}}\arrayrulecolor[HTML]{#2}\cline{#3}}

\setlength{\tabcolsep}{0pt}

\renewcommand*{\arraystretch}{1.5}

\begin{longtable}[c]{|p{1.48in}|p{1.20in}|p{1.20in}|p{1.20in}|p{1.20in}}

\caption{\label{tbl-learning1-results-compare-table}Average score for 10 runs of each algorithm, on each reward function and
each scenario. The standard deviation is shown inside parentheses. } \\

\hhline{>{\arrayrulecolor[HTML]{666666}\global\arrayrulewidth=2pt}->{\arrayrulecolor[HTML]{666666}\global\arrayrulewidth=2pt}->{\arrayrulecolor[HTML]{666666}\global\arrayrulewidth=2pt}->{\arrayrulecolor[HTML]{666666}\global\arrayrulewidth=2pt}->{\arrayrulecolor[HTML]{666666}\global\arrayrulewidth=2pt}-}

\multicolumn{1}{!{\color[HTML]{000000}\vrule width 0pt}>{\raggedright}p{\dimexpr 1.48in+0\tabcolsep+0\arrayrulewidth}}{\textcolor[HTML]{000000}{\fontsize{11}{11}\selectfont{\textbf{RewardFunction}}}} & \multicolumn{1}{!{\color[HTML]{000000}\vrule width 0pt}>{\raggedright}p{\dimexpr 1.2in+0\tabcolsep+0\arrayrulewidth}}{\textcolor[HTML]{000000}{\fontsize{11}{11}\selectfont{\textbf{QSOM}}}} & \multicolumn{1}{!{\color[HTML]{000000}\vrule width 0pt}>{\raggedright}p{\dimexpr 1.2in+0\tabcolsep+0\arrayrulewidth}}{\textcolor[HTML]{000000}{\fontsize{11}{11}\selectfont{\textbf{QDSOM}}}} & \multicolumn{1}{!{\color[HTML]{000000}\vrule width 0pt}>{\raggedright}p{\dimexpr 1.2in+0\tabcolsep+0\arrayrulewidth}}{\textcolor[HTML]{000000}{\fontsize{11}{11}\selectfont{\textbf{DDPG}}}} & \multicolumn{1}{!{\color[HTML]{000000}\vrule width 0pt}>{\raggedright}p{\dimexpr 1.2in+0\tabcolsep+0\arrayrulewidth}!{\color[HTML]{000000}\vrule width 0pt}}{\textcolor[HTML]{000000}{\fontsize{11}{11}\selectfont{\textbf{MADDPG}}}} \\

\hhline{>{\arrayrulecolor[HTML]{666666}\global\arrayrulewidth=2pt}->{\arrayrulecolor[HTML]{666666}\global\arrayrulewidth=2pt}->{\arrayrulecolor[HTML]{666666}\global\arrayrulewidth=2pt}->{\arrayrulecolor[HTML]{666666}\global\arrayrulewidth=2pt}->{\arrayrulecolor[HTML]{666666}\global\arrayrulewidth=2pt}-}\endhead

\multicolumn{5}{!{\color[HTML]{000000}\vrule width 0pt}>{\raggedright}p{\dimexpr 6.28in+8\tabcolsep+4\arrayrulewidth}!{\color[HTML]{000000}\vrule width 0pt}}{\textcolor[HTML]{000000}{\fontsize{10}{10}\selectfont{\textbf{Scenario}}}\textcolor[HTML]{000000}{\fontsize{10}{10}\selectfont{\textbf{:\ }}}\textcolor[HTML]{000000}{\fontsize{10}{10}\selectfont{\textbf{daily\ /\ small}}}} \\

\multicolumn{1}{!{\color[HTML]{000000}\vrule width 0pt}>{\raggedright}p{\dimexpr 1.48in+0\tabcolsep+0\arrayrulewidth}}{\textcolor[HTML]{000000}{\fontsize{10}{10}\selectfont{equity}}} & \multicolumn{1}{!{\color[HTML]{000000}\vrule width 0pt}>{\raggedright}p{\dimexpr 1.2in+0\tabcolsep+0\arrayrulewidth}}{\textcolor[HTML]{000000}{\fontsize{10}{10}\selectfont{1.00\ (+/-\ 0.00\ )}}} & \multicolumn{1}{!{\color[HTML]{000000}\vrule width 0pt}>{\raggedright}p{\dimexpr 1.2in+0\tabcolsep+0\arrayrulewidth}}{\textcolor[HTML]{000000}{\fontsize{10}{10}\selectfont{0.99\ (+/-\ 0.00\ )}}} & \multicolumn{1}{!{\color[HTML]{000000}\vrule width 0pt}>{\raggedright}p{\dimexpr 1.2in+0\tabcolsep+0\arrayrulewidth}}{\textcolor[HTML]{000000}{\fontsize{10}{10}\selectfont{0.99\ (+/-\ 0.01\ )}}} & \multicolumn{1}{!{\color[HTML]{000000}\vrule width 0pt}>{\raggedright}p{\dimexpr 1.2in+0\tabcolsep+0\arrayrulewidth}!{\color[HTML]{000000}\vrule width 0pt}}{\textcolor[HTML]{000000}{\fontsize{10}{10}\selectfont{0.56\ (+/-\ 0.11\ )}}} \\
\end{longtable}

\end{table}

The results presented in Figure~\ref{fig-learning1-results-compare} and
Table~\ref{tbl-learning1-results-compare-table} show that the QSOM
algorithm performs better. We use the Wilcoxon statistical test, which
is the non-parametric equivalent of the well-known T-test, to determine
whether there is a statistically significant difference in the means of
runs' scores between different algorithms. Wilcoxon's test, when used
with the \emph{greater} alternative, assumes as a null hypothesis that
the 2 algorithms have similar means, or that the observed difference is
negligible and only due to chance. The Wilcoxon method returns the
\emph{p-value}, i.e, the likelihood of the null hypothesis being true.
When \(p < \alpha = 0.05\), we say that it is more likely that the null
hypothesis can be refuted, and we assume that the alternative hypothesis
is the correct one. The alternative hypothesis, in this case, is that
the QSOM algorithm obtains better results than its opposing algorithm.
We thus compare algorithms 2-by-2, on each reward function and scenario.

The statistics, presented in Table
Table~\ref{tbl-learning1-compare-stats-qsom}, prove that the QSOM
algorithm statistically outperforms other algorithms, in particular DDPG
and MADDPG, on most scenarii and reward functions, except a few cases,
indicated by the absence of \texttt{*} next to the \emph{p-value}. For
example, DDPG obtains similar scores on the \emph{daily / small
overconsumption} and \emph{multiobj-prod} cases, as well as \emph{daily
/ medium overconsumption}, and \emph{annual / medium overconsumption}.
QDSOM is also quite on par with QSOM on the \emph{daily / small
adaptability1} case. Yet, MADDPG is consistently outperformed by QSOM.

\hypertarget{tbl-learning1-compare-stats-qsom}{}
\begin{table}[H]

\providecommand{\docline}[3]{\noalign{\global\setlength{\arrayrulewidth}{#1}}\arrayrulecolor[HTML]{#2}\cline{#3}}

\setlength{\tabcolsep}{0pt}

\renewcommand*{\arraystretch}{1.5}

\begin{longtable}[c]{|p{1.48in}|p{1.03in}|p{1.03in}|p{1.03in}}

\caption{\label{tbl-learning1-compare-stats-qsom}Comparison of the \emph{QSOM} algorithm with the others, using a
Wilcoxon statistical test, with the `greater' alternative. } \\

\hhline{>{\arrayrulecolor[HTML]{666666}\global\arrayrulewidth=2pt}->{\arrayrulecolor[HTML]{666666}\global\arrayrulewidth=2pt}->{\arrayrulecolor[HTML]{666666}\global\arrayrulewidth=2pt}->{\arrayrulecolor[HTML]{666666}\global\arrayrulewidth=2pt}-}

\multicolumn{1}{!{\color[HTML]{000000}\vrule width 0pt}>{\raggedright}p{\dimexpr 1.48in+0\tabcolsep+0\arrayrulewidth}}{\textcolor[HTML]{000000}{\fontsize{11}{11}\selectfont{\textbf{}}}} & \multicolumn{3}{!{\color[HTML]{000000}\vrule width 0pt}>{\raggedright}p{\dimexpr 3.08in+4\tabcolsep+2\arrayrulewidth}!{\color[HTML]{000000}\vrule width 0pt}}{\textcolor[HTML]{000000}{\fontsize{11}{11}\selectfont{\textbf{Wilcoxon's\ p-value\ (QSOM\ vs\ ...)}}}} \\

\hhline{>{\arrayrulecolor[HTML]{666666}\global\arrayrulewidth=2pt}->{\arrayrulecolor[HTML]{666666}\global\arrayrulewidth=2pt}->{\arrayrulecolor[HTML]{666666}\global\arrayrulewidth=2pt}->{\arrayrulecolor[HTML]{666666}\global\arrayrulewidth=2pt}-}

\multicolumn{1}{!{\color[HTML]{000000}\vrule width 0pt}>{\raggedright}p{\dimexpr 1.48in+0\tabcolsep+0\arrayrulewidth}}{\textcolor[HTML]{000000}{\fontsize{11}{11}\selectfont{\textbf{RewardFunction}}}} & \multicolumn{1}{!{\color[HTML]{000000}\vrule width 0pt}>{\raggedright}p{\dimexpr 1.03in+0\tabcolsep+0\arrayrulewidth}}{\textcolor[HTML]{000000}{\fontsize{11}{11}\selectfont{\textbf{QDSOM}}}} & \multicolumn{1}{!{\color[HTML]{000000}\vrule width 0pt}>{\raggedright}p{\dimexpr 1.03in+0\tabcolsep+0\arrayrulewidth}}{\textcolor[HTML]{000000}{\fontsize{11}{11}\selectfont{\textbf{DDPG}}}} & \multicolumn{1}{!{\color[HTML]{000000}\vrule width 0pt}>{\raggedright}p{\dimexpr 1.03in+0\tabcolsep+0\arrayrulewidth}!{\color[HTML]{000000}\vrule width 0pt}}{\textcolor[HTML]{000000}{\fontsize{11}{11}\selectfont{\textbf{MADDPG}}}} \\

\hhline{>{\arrayrulecolor[HTML]{666666}\global\arrayrulewidth=2pt}->{\arrayrulecolor[HTML]{666666}\global\arrayrulewidth=2pt}->{\arrayrulecolor[HTML]{666666}\global\arrayrulewidth=2pt}->{\arrayrulecolor[HTML]{666666}\global\arrayrulewidth=2pt}-}\endhead

\multicolumn{4}{!{\color[HTML]{000000}\vrule width 0pt}>{\raggedright}p{\dimexpr 4.56in+6\tabcolsep+3\arrayrulewidth}!{\color[HTML]{000000}\vrule width 0pt}}{\textcolor[HTML]{000000}{\fontsize{10}{10}\selectfont{\textbf{Scenario}}}\textcolor[HTML]{000000}{\fontsize{10}{10}\selectfont{\textbf{:\ }}}\textcolor[HTML]{000000}{\fontsize{10}{10}\selectfont{\textbf{daily\ /\ small}}}} \\

\multicolumn{1}{!{\color[HTML]{000000}\vrule width 0pt}>{\raggedright}p{\dimexpr 1.48in+0\tabcolsep+0\arrayrulewidth}}{\textcolor[HTML]{000000}{\fontsize{10}{10}\selectfont{equity}}} & \multicolumn{1}{!{\color[HTML]{000000}\vrule width 0pt}>{\raggedright}p{\dimexpr 1.03in+0\tabcolsep+0\arrayrulewidth}}{\textcolor[HTML]{000000}{\fontsize{10}{10}\selectfont{5.41e-06***}}} & \multicolumn{1}{!{\color[HTML]{000000}\vrule width 0pt}>{\raggedright}p{\dimexpr 1.03in+0\tabcolsep+0\arrayrulewidth}}{\textcolor[HTML]{000000}{\fontsize{10}{10}\selectfont{5.41e-06***}}} & \multicolumn{1}{!{\color[HTML]{000000}\vrule width 0pt}>{\raggedright}p{\dimexpr 1.03in+0\tabcolsep+0\arrayrulewidth}!{\color[HTML]{000000}\vrule width 0pt}}{\textcolor[HTML]{000000}{\fontsize{10}{10}\selectfont{5.41e-06***}}} \\

\multicolumn{1}{!{\color[HTML]{000000}\vrule width 0pt}>{\raggedright}p{\dimexpr 1.48in+0\tabcolsep+0\arrayrulewidth}}{\textcolor[HTML]{000000}{\fontsize{10}{10}\selectfont{overconsumption}}} & \multicolumn{1}{!{\color[HTML]{000000}\vrule width 0pt}>{\raggedright}p{\dimexpr 1.03in+0\tabcolsep+0\arrayrulewidth}}{\textcolor[HTML]{000000}{\fontsize{10}{10}\selectfont{0.005748\ **}}} & \multicolumn{1}{!{\color[HTML]{000000}\vrule width 0pt}>{\raggedright}p{\dimexpr 1.03in+0\tabcolsep+0\arrayrulewidth}}{\textcolor[HTML]{000000}{\fontsize{10}{10}\selectfont{0.289371\ \ \ }}} & \multicolumn{1}{!{\color[HTML]{000000}\vrule width 0pt}>{\raggedright}p{\dimexpr 1.03in+0\tabcolsep+0\arrayrulewidth}!{\color[HTML]{000000}\vrule width 0pt}}{\textcolor[HTML]{000000}{\fontsize{10}{10}\selectfont{0.000103***}}} \\
\end{longtable}

\end{table}

\hypertarget{tbl-learning1-compare-stats-qdsom}{}
\begin{table}[H]

\providecommand{\docline}[3]{\noalign{\global\setlength{\arrayrulewidth}{#1}}\arrayrulecolor[HTML]{#2}\cline{#3}}

\setlength{\tabcolsep}{0pt}

\renewcommand*{\arraystretch}{1.5}

\begin{longtable}[c]{|p{1.48in}|p{0.75in}|p{1.03in}|p{1.03in}}

\caption{\label{tbl-learning1-compare-stats-qdsom}Comparison of the \emph{QDSOM} algorithm with the others, using a
Wilcoxon statistical test, with the `greater' alternative. } \\

\hhline{>{\arrayrulecolor[HTML]{666666}\global\arrayrulewidth=2pt}->{\arrayrulecolor[HTML]{666666}\global\arrayrulewidth=2pt}->{\arrayrulecolor[HTML]{666666}\global\arrayrulewidth=2pt}->{\arrayrulecolor[HTML]{666666}\global\arrayrulewidth=2pt}-}

\multicolumn{1}{!{\color[HTML]{000000}\vrule width 0pt}>{\raggedright}p{\dimexpr 1.48in+0\tabcolsep+0\arrayrulewidth}}{\textcolor[HTML]{000000}{\fontsize{11}{11}\selectfont{\textbf{}}}} & \multicolumn{3}{!{\color[HTML]{000000}\vrule width 0pt}>{\raggedright}p{\dimexpr 2.81in+4\tabcolsep+2\arrayrulewidth}!{\color[HTML]{000000}\vrule width 0pt}}{\textcolor[HTML]{000000}{\fontsize{11}{11}\selectfont{\textbf{Wilcoxon's\ p-value\ (QDSOM\ vs\ ...)}}}} \\

\hhline{>{\arrayrulecolor[HTML]{666666}\global\arrayrulewidth=2pt}->{\arrayrulecolor[HTML]{666666}\global\arrayrulewidth=2pt}->{\arrayrulecolor[HTML]{666666}\global\arrayrulewidth=2pt}->{\arrayrulecolor[HTML]{666666}\global\arrayrulewidth=2pt}-}

\multicolumn{1}{!{\color[HTML]{000000}\vrule width 0pt}>{\raggedright}p{\dimexpr 1.48in+0\tabcolsep+0\arrayrulewidth}}{\textcolor[HTML]{000000}{\fontsize{11}{11}\selectfont{\textbf{RewardFunction}}}} & \multicolumn{1}{!{\color[HTML]{000000}\vrule width 0pt}>{\raggedright}p{\dimexpr 0.75in+0\tabcolsep+0\arrayrulewidth}}{\textcolor[HTML]{000000}{\fontsize{11}{11}\selectfont{\textbf{QSOM}}}} & \multicolumn{1}{!{\color[HTML]{000000}\vrule width 0pt}>{\raggedright}p{\dimexpr 1.03in+0\tabcolsep+0\arrayrulewidth}}{\textcolor[HTML]{000000}{\fontsize{11}{11}\selectfont{\textbf{DDPG}}}} & \multicolumn{1}{!{\color[HTML]{000000}\vrule width 0pt}>{\raggedright}p{\dimexpr 1.03in+0\tabcolsep+0\arrayrulewidth}!{\color[HTML]{000000}\vrule width 0pt}}{\textcolor[HTML]{000000}{\fontsize{11}{11}\selectfont{\textbf{MADDPG}}}} \\

\hhline{>{\arrayrulecolor[HTML]{666666}\global\arrayrulewidth=2pt}->{\arrayrulecolor[HTML]{666666}\global\arrayrulewidth=2pt}->{\arrayrulecolor[HTML]{666666}\global\arrayrulewidth=2pt}->{\arrayrulecolor[HTML]{666666}\global\arrayrulewidth=2pt}-}\endhead

\multicolumn{4}{!{\color[HTML]{000000}\vrule width 0pt}>{\raggedright}p{\dimexpr 4.29in+6\tabcolsep+3\arrayrulewidth}!{\color[HTML]{000000}\vrule width 0pt}}{\textcolor[HTML]{000000}{\fontsize{10}{10}\selectfont{\textbf{Scenario}}}\textcolor[HTML]{000000}{\fontsize{10}{10}\selectfont{\textbf{:\ }}}\textcolor[HTML]{000000}{\fontsize{10}{10}\selectfont{\textbf{daily\ /\ small}}}} \\

\multicolumn{1}{!{\color[HTML]{000000}\vrule width 0pt}>{\raggedright}p{\dimexpr 1.48in+0\tabcolsep+0\arrayrulewidth}}{\textcolor[HTML]{000000}{\fontsize{10}{10}\selectfont{equity}}} & \multicolumn{1}{!{\color[HTML]{000000}\vrule width 0pt}>{\raggedright}p{\dimexpr 0.75in+0\tabcolsep+0\arrayrulewidth}}{\textcolor[HTML]{000000}{\fontsize{10}{10}\selectfont{1.000\ \ \ }}} & \multicolumn{1}{!{\color[HTML]{000000}\vrule width 0pt}>{\raggedright}p{\dimexpr 1.03in+0\tabcolsep+0\arrayrulewidth}}{\textcolor[HTML]{000000}{\fontsize{10}{10}\selectfont{0.999475\ \ \ }}} & \multicolumn{1}{!{\color[HTML]{000000}\vrule width 0pt}>{\raggedright}p{\dimexpr 1.03in+0\tabcolsep+0\arrayrulewidth}!{\color[HTML]{000000}\vrule width 0pt}}{\textcolor[HTML]{000000}{\fontsize{10}{10}\selectfont{5.41e-06***}}} \\

\multicolumn{1}{!{\color[HTML]{000000}\vrule width 0pt}>{\raggedright}p{\dimexpr 1.48in+0\tabcolsep+0\arrayrulewidth}}{\textcolor[HTML]{000000}{\fontsize{10}{10}\selectfont{overconsumption}}} & \multicolumn{1}{!{\color[HTML]{000000}\vrule width 0pt}>{\raggedright}p{\dimexpr 0.75in+0\tabcolsep+0\arrayrulewidth}}{\textcolor[HTML]{000000}{\fontsize{10}{10}\selectfont{0.996\ \ \ }}} & \multicolumn{1}{!{\color[HTML]{000000}\vrule width 0pt}>{\raggedright}p{\dimexpr 1.03in+0\tabcolsep+0\arrayrulewidth}}{\textcolor[HTML]{000000}{\fontsize{10}{10}\selectfont{0.995535\ \ \ }}} & \multicolumn{1}{!{\color[HTML]{000000}\vrule width 0pt}>{\raggedright}p{\dimexpr 1.03in+0\tabcolsep+0\arrayrulewidth}!{\color[HTML]{000000}\vrule width 0pt}}{\textcolor[HTML]{000000}{\fontsize{10}{10}\selectfont{0.000752***}}} \\
\end{longtable}

\end{table}

\begin{figure}

{\centering \includegraphics{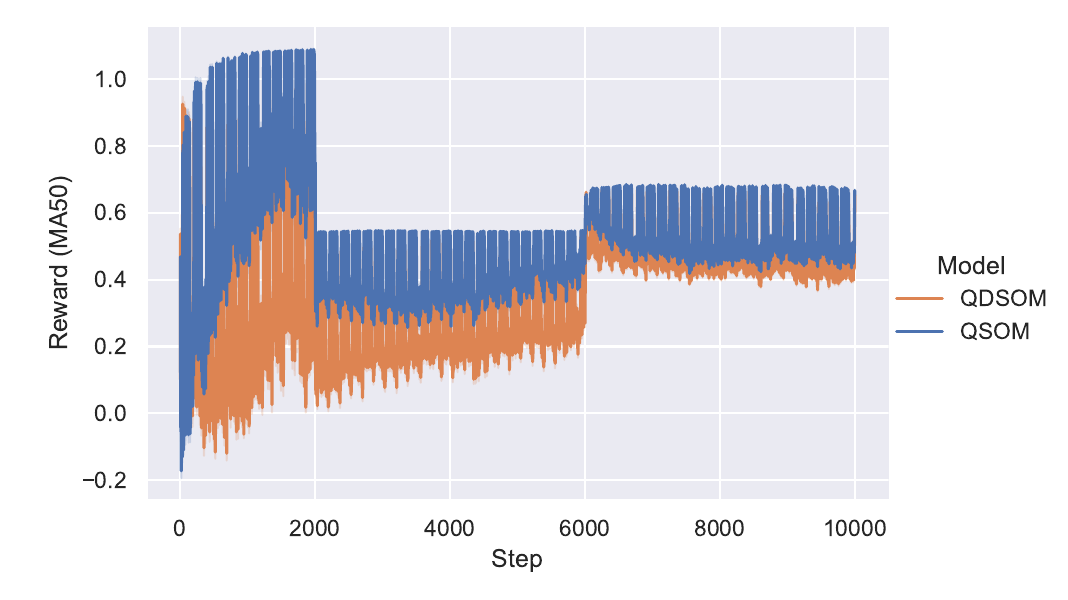}

}

\caption{\label{fig-learning1-agents-rewards}Agents Rewards}

\end{figure}

Figure~\ref{fig-learning1-agents-rewards} shows the evolution of
individual rewards received by agents over the time steps, in the
\emph{annual / small} scenario, using the \emph{adaptability2} reward
function. We chose to focus on this combination of scenario and reward
function as they are, arguably, the most interesting. \emph{Daily}
scenarii are perhaps too easy for the agents as they do not include as
many variations as the \emph{annual}; additionally, \emph{small}
scenarios are easier to visualize and explore, as they contain fewer
agents than \emph{medium} scenarios. Finally, the \emph{adaptability2}
is retained for the same arguments that made us choose it for the
hyperparameters search. We show a moving average of the rewards in order
to erase the small and local variations to highlight the larger trend of
the rewards' evolution.

We can see from the results that the \emph{small} scenarii seem to yield
a slightly better score than the \emph{medium} scenarii. Thus, agents
are impacted by the increased number of other agents, and have
difficulties learning the whole environment dynamics. Still, the results
reported for the \emph{medium} scenarii are near the \emph{small}
results, and very close to \(1\). Even though there is indeed an effect
of the environment size on the score, this hints towards the scalability
of our approach, as the agents managed to learn a ``good'' behaviour
that yields high rewards.

\hypertarget{sec-discussion}{%
\section{Discussion}\label{sec-discussion}}

In this article, we presented two new reinforcement learning algorithms,
QSOM and QDSOM.

We recall that the principal important aspects and limitations
identified in the State of the Art were the following:

\begin{itemize}
\tightlist
\item
  Using continuous and multi-dimensional domains to improve the
  environment's richness.
\item
  Continuously learning and adapting to changes in the environment,
  including in the reward function, i.e., the structure that encodes and
  captures the ethical considerations that agents should learn to
  exhibit.
\item
  Learning in a multi-agent setting, by taking into account the
  difficulties posed by the presence of other agents.
\end{itemize}

The continuous and multi-dimensional aspect was solved by design, thanks
to the SOMs and DSOMs that we use in our algorithms. They learn to
handle the complex observations and actions domains, while
advantageously offering a discrete representation that can be leveraged
with the Q-Table, permitting a modular approach. This modular approach,
and the use of Q-Tables, allow for example to compare different actions,
which is not always possible in end-to-end Deep Neural Networks.

The continuous adaptation was also handled by our design choices,
notably by disabling traditional convergence mechanisms. The use of
(D)SOMs also help, as the representation may shift over time by moving
the neurons. Additionally, our experiments highlight the ability of our
algorithms to adapt, especially when compared to other algorithms,
through the specific \emph{adaptability1} and \emph{adaptability2}
functions.

Finally, challenges raised by the multi-agent aspect were partially
answered by the use of Difference Rewards to create the reward
functions. On the other hand, the agents themselves have no specific
mechanism that help them learn a behaviour while taking account of the
other agents in the shared environment, e.g., contrary to Centralized
Training algorithms such as MADDPG. Nevertheless, our algorithms managed
to perform better than MADDPG on the proposed scenarii and reward
functions, which means that this limitation is not crippling.

Our algorithms still suffer from a few limitations that we highlight
here.

\begin{itemize}
\item
  As we already mentioned, the multi-agent aspect could be improved, for
  example by adding communication mechanisms between agents. Indeed, by
  being able to communicate, agents could coordinate their actions so
  that the joint-action could be even better. Let us assume that an
  agent, which approximately learned the environment dynamics, believes
  that there is not much consumption at 3AM, and chooses the strategy of
  replenishing its battery at this moment, so as to have a minimal
  impact on the grid. Another agent may, at some point, face an urgent
  situation that requires it to consume exceptionally at 3AM this day.
  Without coordination, the 2 agents will both consume an import amount
  of energy at the same time, thus impacting the grid and potentially
  over-consuming. On the other hand, if the agents could communicate,
  the second one may inform other agents of its urgency. The first one
  would perhaps choose to consume only at 4AM, or they would both
  negotiate an amount of energy to share, in the end proposing a better
  joint-action than the uncoordinated sum of their individual actions.
  However, such communication should be carefully designed in a
  privacy-respectful manner.
\item
  The algorithms have not been tested on other (baseline) environments.
  This limits their results and promises: it might happen that their
  success, compared to DDPG and MADDPG especially, is due to some
  specificities of the Smart Grid environment. In particular, recent
  Deep Reinforcement Learning algorithms target use-cases with huge
  state-action spaces, e.g., taking input directly from the screen
  (pixels), or emitting physical actions concerning dozens of joints.
  Although our use-case used more dimensions (11 for states and 6 for
  actions) than Smith's original experiments, it does not compare to
  such environments. The performance of QSOM and QDSOM on them thus
  remains uncertain.
\end{itemize}

\newpage{}

\hypertarget{references}{%
\section*{References}\label{references}}
\addcontentsline{toc}{section}{References}

\hypertarget{refs}{}
\begin{CSLReferences}{1}{0}
\leavevmode\vadjust pre{\hypertarget{ref-allen2005artificial}{}}%
Allen, Colin, Iva Smit, and Wendell Wallach. 2005. {``Artificial
Morality: Top-down, Bottom-up, and Hybrid Approaches.''} \emph{Ethics
and Information Technology} 7 (3): 149--55.

\leavevmode\vadjust pre{\hypertarget{ref-andersonMachineEthics2011}{}}%
Anderson, Michael, and Susan Leigh Anderson. 2011. \emph{Machine
{Ethics}}. {Cambridge University Press}.
\url{https://books.google.com?id=N4IF2p4w7uwC}.

\leavevmode\vadjust pre{\hypertarget{ref-anderson2018value}{}}%
Anderson, Michael, Susan Leigh Anderson, and Vincent Berenz. 2018. {``A
Value-Driven Eldercare Robot: Virtual and Physical Instantiations of a
Case-Supported Principle-Based Behavior Paradigm.''} \emph{Proceedings
of the IEEE} 107 (3): 526--40.

\leavevmode\vadjust pre{\hypertarget{ref-bellman}{}}%
Bellman, Richard. 1966. {``Dynamic Programming.''} \emph{Science} 153
(3731): 34--37.

\leavevmode\vadjust pre{\hypertarget{ref-bernstein2002complexity}{}}%
Bernstein, Daniel S, Robert Givan, Neil Immerman, and Shlomo
Zilberstein. 2002. {``The Complexity of Decentralized Control of Markov
Decision Processes.''} \emph{Mathematics of Operations Research} 27 (4):
819--40.

\leavevmode\vadjust pre{\hypertarget{ref-bremner2019proactiveTransparentVerifiable}{}}%
Bremner, Paul, Louise A. Dennis, Michael Fisher, and Alan F. Winfield.
2019. {``On {Proactive}, {Transparent}, and {Verifiable Ethical
Reasoning} for {Robots}.''} \emph{Proceedings of the IEEE} 107 (3):
541--61. \url{https://doi.org/10.1109/JPROC.2019.2898267}.

\leavevmode\vadjust pre{\hypertarget{ref-cointe2016ethicalJudgmentAgents}{}}%
Cointe, Nicolas, Grégory Bonnet, and Olivier Boissier. 2016. {``Ethical
Judgment of Agents' Behaviors in Multi-Agent Systems.''} In
\emph{Proceedings of the 2016 International Conference on Autonomous
Agents \& Multiagent Systems}, 1106--14. AAMAS '16. Richland, SC:
International Foundation for Autonomous Agents; Multiagent Systems.

\leavevmode\vadjust pre{\hypertarget{ref-dignum2019responsible}{}}%
Dignum, Virginia. 2019. \emph{Responsible {Artificial Intelligence}:
{How} to {Develop} and {Use AI} in a {Responsible Way}}. Artificial
{Intelligence}: {Foundations}, {Theory}, and {Algorithms}. {Springer
International Publishing}.
\url{https://doi.org/10.1007/978-3-030-30371-6}.

\leavevmode\vadjust pre{\hypertarget{ref-foot1967problem}{}}%
Foot, Philippa. 1967. {``The Problem of Abortion and the Doctrine of the
Double Effect.''} \emph{Oxford Review} 5.

\leavevmode\vadjust pre{\hypertarget{ref-gini1936measure}{}}%
Gini, Corrado. 1936. {``On the Measure of Concentration with Special
Reference to Income and Statistics.''} \emph{Colorado College
Publication, General Series} 208 (1): 73--79.

\leavevmode\vadjust pre{\hypertarget{ref-haas2020moralGridworlds}{}}%
Haas, Julia. 2020. {``Moral {Gridworlds}: {A Theoretical Proposal} for
{Modeling Artificial Moral Cognition}.''} \emph{Minds and Machines},
April. \url{https://doi.org/10.1007/s11023-020-09524-9}.

\leavevmode\vadjust pre{\hypertarget{ref-honarvar2009artificial}{}}%
Honarvar, Ali Reza, and Nasser Ghasem-Aghaee. 2009. {``An Artificial
Neural Network Approach for Creating an Ethical Artificial Agent.''} In
\emph{2009 IEEE International Symposium on Computational Intelligence in
Robotics and Automation-(CIRA)}, 290--95. IEEE.

\leavevmode\vadjust pre{\hypertarget{ref-kohonen1990self}{}}%
Kohonen, Teuvo. 1990. {``The Self-Organizing Map.''} \emph{Proceedings
of the IEEE} 78 (9): 1464--80.

\leavevmode\vadjust pre{\hypertarget{ref-lillicrap2015continuous}{}}%
Lillicrap, Timothy P, Jonathan J Hunt, Alexander Pritzel, Nicolas Heess,
Tom Erez, Yuval Tassa, David Silver, and Daan Wierstra. 2015.
{``Continuous Control with Deep Reinforcement Learning.''} \emph{arXiv
Preprint arXiv:1509.02971}.

\leavevmode\vadjust pre{\hypertarget{ref-lowe2017multi}{}}%
Lowe, Ryan, Yi Wu, Aviv Tamar, Jean Harb, Pieter Abbeel, and Igor
Mordatch. 2017. {``Multi-Agent Actor-Critic for Mixed
Cooperative-Competitive Environments.''} \emph{Neural Information
Processing Systems (NIPS)}.

\leavevmode\vadjust pre{\hypertarget{ref-marcus2018deep}{}}%
Marcus, Gary. 2018. {``Deep Learning: A Critical Appraisal.''}
\emph{arXiv Preprint arXiv:1801.00631}.

\leavevmode\vadjust pre{\hypertarget{ref-moor2009four}{}}%
Moor, James. 2009. {``Four Kinds of Ethical Robots.''} \emph{Philosophy
Now} 72: 12--14.

\leavevmode\vadjust pre{\hypertarget{ref-murukannaiah2020new}{}}%
Murukannaiah, Pradeep K, Nirav Ajmeri, Catholijn M Jonker, and Munindar
P Singh. 2020. {``New Foundations of Ethical Multiagent Systems.''} In
\emph{Proceedings of the 19th International Conference on Autonomous
Agents and MultiAgent Systems}, 1706--10.

\leavevmode\vadjust pre{\hypertarget{ref-nallur2020landscape}{}}%
Nallur, Vivek. 2020. {``Landscape of Machine Implemented Ethics.''}
\emph{Science and Engineering Ethics} 26 (5): 2381--99.

\leavevmode\vadjust pre{\hypertarget{ref-openei2014dataset}{}}%
Ong, Sean, and Nathan Clark. 2014. {``Commercial and Residential Hourly
Load Profiles for All TMY3 Locations in the United States
{[}Dataset{]}.''} \url{https://doi.org/10.25984/1788456}.

\leavevmode\vadjust pre{\hypertarget{ref-panait2005cooperativeMultiAgentLearning}{}}%
Panait, Liviu, and Sean Luke. 2005. {``Cooperative {Multi-Agent
Learning}: {The State} of the {Art}.''} \emph{Autonomous Agents and
Multi-Agent Systems} 11 (3): 387--434.
\url{https://doi.org/10.1007/s10458-005-2631-2}.

\leavevmode\vadjust pre{\hypertarget{ref-rougier2011dynamic}{}}%
Rougier, Nicolas, and Yann Boniface. 2011. {``Dynamic Self-Organising
Map.''} \emph{Neurocomputing} 74 (11): 1840--47.

\leavevmode\vadjust pre{\hypertarget{ref-smith2002applications}{}}%
Smith, Andrew James. 2002a. {``Applications of the Self-Organising Map
to Reinforcement Learning.''} \emph{Neural Networks} 15 (8-9): 1107--24.

\leavevmode\vadjust pre{\hypertarget{ref-smith2002dynamic}{}}%
---------. 2002b. {``Dynamic Generalisation of Continuous Action Spaces
in Reinforcement Learning: A Neurally Inspired Approach.''} PhD thesis,
University of Edinburgh.

\leavevmode\vadjust pre{\hypertarget{ref-sutton2018reinforcement}{}}%
Sutton, Richard S, and Andrew G Barto. 2018. \emph{Reinforcement
Learning: An Introduction}. MIT press.

\leavevmode\vadjust pre{\hypertarget{ref-watkins1992qlearning}{}}%
Watkins, Christopher JCH, and Peter Dayan. 1992. {``Q-Learning.''}
\emph{Machine Learning} 8 (3): 279--92.

\leavevmode\vadjust pre{\hypertarget{ref-russell2015valueAlignment}{}}%
World Economic Forum. 2015. {``Value {Alignment} \textbar{} {Stuart
Russell}.''} Youtube. 2015.
\url{https://www.youtube.com/watch?v=WvmeTaFc_Qw}.

\leavevmode\vadjust pre{\hypertarget{ref-wu2018low}{}}%
Wu, Yueh-Hua, and Shou-De Lin. 2018. {``A Low-Cost Ethics Shaping
Approach for Designing Reinforcement Learning Agents.''} In
\emph{Proceedings of the AAAI Conference on Artificial Intelligence}.
Vol. 32. 1.

\leavevmode\vadjust pre{\hypertarget{ref-yliniemi2014Multiobjective}{}}%
Yliniemi, Logan, and Kagan Tumer. 2014. {``Multi-Objective {Multiagent
Credit Assignment Through Difference Rewards} in {Reinforcement
Learning}.''} In \emph{Simulated {Evolution} and {Learning}}, edited by
Grant Dick, Will N. Browne, Peter Whigham, Mengjie Zhang, Lam Thu Bui,
Hisao Ishibuchi, Yaochu Jin, et al., 407--18. Lecture {Notes} in
{Computer Science}. {Springer International Publishing}.

\leavevmode\vadjust pre{\hypertarget{ref-yu2018building}{}}%
Yu, Han, Zhiqi Shen, Chunyan Miao, Cyril Leung, Victor R Lesser, and
Qiang Yang. 2018. {``Building Ethics into Artificial Intelligence.''} In
\emph{Proceedings of the 27th International Joint Conference on
Artificial Intelligence}, 5527--33.

\end{CSLReferences}

\end{document}